\documentclass[times, review, 10pt]{elsarticle}

\usepackage{amssymb}
\usepackage{amsmath}

\usepackage{natbib}   
\usepackage{graphicx}%
\usepackage{multirow}%
\usepackage{amsmath,amssymb,amsfonts}%
\usepackage{amsthm}%
\usepackage{mathrsfs}%
\usepackage{xcolor}%
\usepackage{textcomp}%
\usepackage{manyfoot}%
\usepackage{booktabs}%
\usepackage{algorithm}%
\usepackage{algorithmicx}%
\usepackage{algpseudocode}%
\usepackage{listings}%
\usepackage{xspace}
\usepackage{bm}
\usepackage{bbding}
\usepackage{tabu}
\usepackage{url}
\usepackage[hidelinks]{hyperref}

\makeatletter
\DeclareRobustCommand\onedot{\futurelet\@let@token\@onedot}
\def\@onedot{\ifx\@let@token.\else.\null\fi\xspace}

\def\eg{\emph{e.g}\onedot} 
\def\ie{\emph{i.e}\onedot}

\def\etal{\emph{et al}\onedot}
\makeatother

\definecolor{baseline}{rgb}{0.8, 0.8, 0.8}

\journal{Pattern Recognition}

\begin{document}

\begin{frontmatter}



\title{SelfDRSC++: Self-supervised dual reversed rolling shutter correction via video interpolation}



\author[1]{Wei Shang}  


\affiliation[1]{organization={Singapore Management University},
	country={Singapore}}

\author[2]{Dongwei Ren \corref{cor1}}   

\author[3]{Wanying Zhang}   

\affiliation[2]{organization={Tianjin University},
	country={China}}

\author[2]{Qilong Wang}  

\affiliation[3]{organization={Harbin Institute of Technology},
	country={China}}

%

\author[4]{Pengfei Zhu}  

\affiliation[4]{organization={Southeast University},
	country={China}}

\author[3]{Wangmeng Zuo}  

\cortext[cor1]{Corresponding author}
\begin{abstract}
Modern consumer cameras often use rolling shutter, capturing scenes row-by-row and causing distortion in dynamic scenes. Existing correction methods rely on supervised learning with high-frame-rate global shutter images as ground truth. We propose SelfDRSC++, a self-supervised framework for RS distortion correction {from simultaneously captured top-to-bottom and bottom-to-top RS images}. A lightweight network with a bidirectional correlation matching block jointly optimizes optical flows and corrected RS features, improving performance with fewer parameters. 
A self-supervised strategy enforces {a physically constrained RS--GS--RS cycle} between input and reconstructed dual reversed RS images. RS reconstruction is formulated as a specialized video frame interpolation task, enabling feasible one-stage training. 
{Extensive experiments on synthetic and real-world data show that SelfDRSC++ achieves competitive quantitative performance, improves perceptual quality, and produces high-frame-rate GS sequences with better temporal consistency.}
\end{abstract}




\end{frontmatter}



\section{Introduction}

\begin{figure*}[!t]\footnotesize
	\centering
	\includegraphics[width=0.70\linewidth]{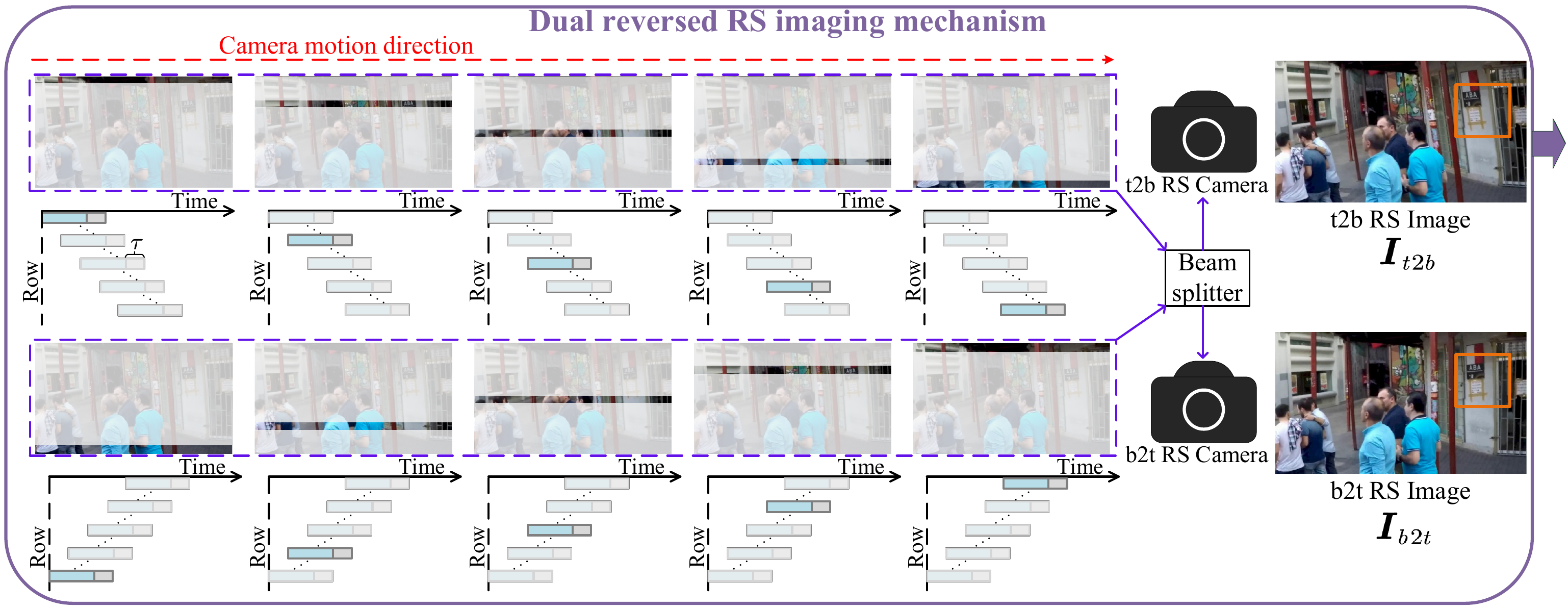}   
	\includegraphics[width=0.29\linewidth]{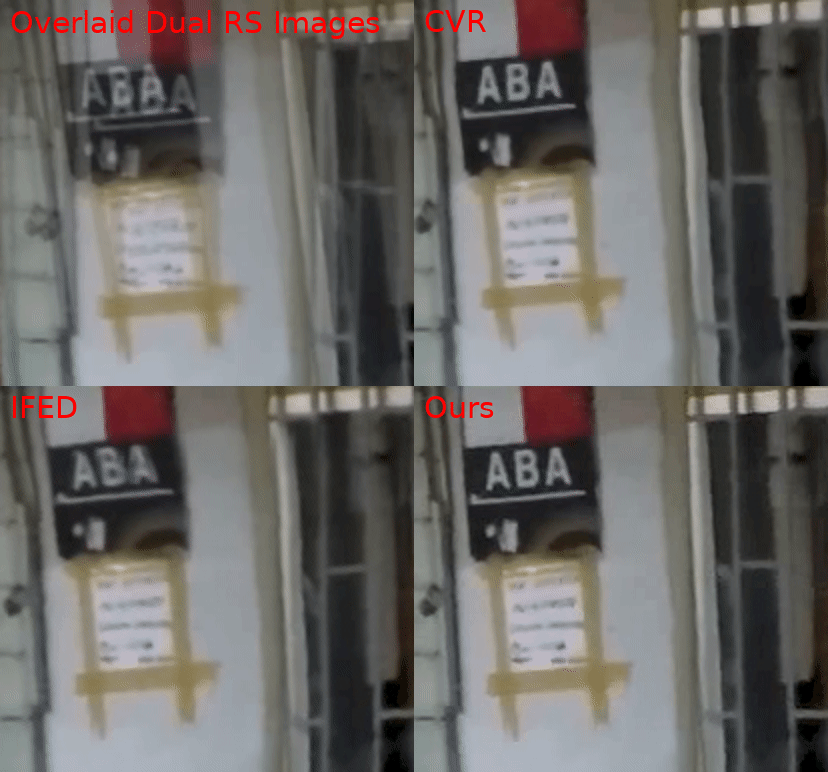}\\
	\vspace{-3mm}
	\caption{
		Illustration of capturing dual RS images with reversed scanning directions, \ie, top-to-bottom ($\bm{I}_{t2b}$) and bottom-to-top ($\bm{I}_{b2t}$). 
		{This setting provides complementary scanline-time observations for recovering latent high-frame-rate GS frames.}
		In comparison to state-of-the-art supervised RS correction methods CVR \cite{Fan_2022_CVPR} and IFED \cite{zhong2022bringing}, our SelfDRSC++ can generate high framerate GS videos with finer textures and better temporal consistency.
	}
	\vspace{-3mm}
	\label{fig:dual RS}
\end{figure*}
Recent years have witnessed an increasing demand for imaging sensors, due to the widespread applications of digital cameras and smartphones. 
Although the Charge-Coupled Device (CCD) has been the dominant technology for imaging sensors, 
{Complementary Metal-Oxide Semiconductor (CMOS) sensors have recently become a popular alternative in modern consumer cameras due to their advantages, such as easy integration with image processing pipelines and communication circuits, as well as low power consumption~\cite{litwiller2001ccd}.}
In CMOS sensors, rolling shutter (RS) scanning mechanism is generally deployed to capture images, \ie, each row of CMOS array is exposed in the sequential time, which is different from CCD with global shutter (GS) scanning at one instant.  
Therefore, RS images suffer from distortion when capturing dynamic scenes, {which not only affects human visual perception but can also degrade, or even break, downstream computer vision tasks~\cite{zhang2017synthetic, munoz2019spm}.}

To correct RS distortion, pioneering works usually reconstruct GS images from a single RS image \cite{zhuang2019learning} or multiple consecutive RS images \cite{liu2020deep,fan2024learning}, where the latter ones usually have better performance. 
But the consecutive RS image setting is ambiguous \cite{zhong2022bringing}, \eg, two RS cameras, moving horizontally at the same speed but with different readout time, can produce the same RS images. 
To address this ambiguity, a new RS acquisition setting, \ie, dual RS images with reversed scanning directions (see Fig. \ref{fig:dual RS}), is proposed in \cite{albl2020two,zhong2022bringing}. {Different from many video-based RS correction methods that use two consecutive same-direction RS frames, this setting still uses two RS observations but changes the acquisition pattern to simultaneous opposite-direction scanning, thereby providing complementary scanline-time cues.}
%
%
Nevertheless, both of them adopt a fully supervised learning manner, \ie, ground-truth GS images are required for supervised learning RS correction networks. 
%
%
However, it is not easy to collect real-world training samples, while synthetic training samples yield poor generalization ability when handling real-world RS cases.   
%
%

%
In this paper, we address the more challenging and practical task, \ie, self-supervised learning to recover a high-frame-rate GS video from dual reversed RS images.
As shown in Fig. \ref{fig:selfsup}, our proposed SelfDRSC++ consists of two key modules, \ie, a dual reversed RS correction (DRSC) network $\mathcal{F}$ for correcting RS distortion and an RS reconstruction module $\mathcal{W}$ to enable feasible self-supervised learning. 
{Unlike prior works that treat RS correction as a monolithic warping or flow-estimation problem, we reformulate the task as row-wise temporal interpolation governed by the differentiable distorted time map and time displacement -- representations that are physically grounded in readout timing.}
For the DRSC network design in Fig. \ref{fig:framework}, {we successfully apply the core modules and strategies from existing VFI to the RS correction task, enabling the network to jointly optimize optical flow and RS features. Compared with IFED~\cite{zhong2022bringing}, it reduces parameters by 90\% while achieving lower LPIPS.} 
%
%
Our self-supervised learning strategy is guided by the principle that the DRSC network's predicted latent GS images can be utilized to reconstruct dual RS images with reversed distortion, thereby enabling the training of the DRSC network by enforcing {a physically constrained RS--GS--RS cycle} between input and reconstructed RS images.
%
%
To achieve this, we propose a VFI-based RS reconstruction module (as depicted in Fig.~\ref{fig:RS gen}), which considers RS reconstruction as a specialized case of video frame interpolation by using a distorted time map to substitute distance indexing in VFI method~\cite{zhong2023clearer}. 
{In this formulation, the distorted time map specifies the row-wise interpolation time induced by RS readout, making GS-to-RS reconstruction a time-indexed VFI problem rather than a generic image synthesis process.}
Finally, we employ a self-supervised loss function to train the DRSC network. 
During the training process, an intermediate GS image at arbitrary RS scanning time is predicted, corresponding to which another set of dual RS images can be reconstructed in our VFI-based RS reconstruction module. 
Consequently, the predicted GS images at intermediate scanning time in SelfDRSC++ can also be supervised, making the learned DRSC network be able to generate high framerate GS videos.

Although any ground-truth GS images are not used in the training phase of our SelfDRSC++, it still achieves comparable performance on synthetic and real-world datasets, in comparison to state-of-the-art supervised RS correction methods. 
This work is previously presented as a conference paper \cite{shang2023self}, where the DRSC network is borrowed from IFED~\cite{zhong2022bringing} and the self-supervised training is conducted in two stages to address boundary artifacts in corrected results. 
In this manuscript, our major improvements include
(\emph{i}) a more compact and lightweight DRSC network that can improve correction performance with reduced network parameters, (\emph{ii}) a one-stage self-supervised training strategy that is able to effectively train DRSC network, and (\emph{iii}) inclusion of more state-of-the-art methods such as JAM \cite{fan2023joint}, LBCNet \cite{fan2024learning}, Uni-SoftSplat~\cite{fan2024unified}, V-DRSC \cite{qu2023fast}, SelfRSSplat~\cite{fan2024self}, {SelfUnroll \cite{lin2025self}, and Self-EvRSVFI \cite{lu2025self}} for experimental comparison. 
To sum up, the contributions of this work are three-fold:
\begin{itemize}
	\item {We establish a one-stage self-supervised RS--GS--RS learning formulation for dual reversed RS correction. By requiring the predicted latent GS frames to reconstruct both top-to-bottom and bottom-to-top RS observations, the proposed formulation provides physically constrained supervision.}
	
	\item {We formulate GS-to-RS reconstruction as a row-wise time-indexed VFI problem driven by distorted time maps. This design provides a differentiable surrogate of the RS image formation process and enables stable self-supervised training while alleviating boundary artifacts caused by flow-based RS reconstruction.}
	
	\item 
	We present a compact and lightweight RS correction network, {which adapts bidirectional correlation matching and multi-field reconstruction to time-conditioned RS correction}, and which significantly surpasses state-of-the-art methods in terms of correction textures for dealing with real-world RS cases.
	
\end{itemize}

\section{Related work}\label{sec:related}

\subsection{Rolling shutter distortion correction}
With the rising demand for RS cameras, RS distortion corrections have received widespread attention. 
Existing works on RS correction generally fall into two categories: single-image-based~\cite{zhuang2019learning} and multi-frame-based~\cite{Fan_2021_ICCV,fan2023joint,qu2023fast} methods. 
It is an ill-posed problem to correct RS distortion from a single image, and its performance is usually inferior. 
For multi-frame-based methods, can be further divided into generating one specific image and generating a video sequence.
For the former, Liu \etal~\cite{liu2020deep} proposed an end-to-end model, which warped features of RS images to a GS image by a special forward warping block.
For the latter, Fan \etal~\cite{Fan_2021_ICCV} designed a deep learning framework, which utilized the underlying spatio-temperal geometric relationships for generating a latent GS image sequence. Then Fan \etal~\cite{Fan_2022_CVPR} further proposed a context-aware model for solving complex occlusions and object-specific motion artifacts.
Recently, Zhong \etal~\cite{zhong2022bringing} proposed an end-to-end method IFED, and it can extract a GS sequence grounded on the symmetric and complementary nature of dual RS images with reversed distortion.
Fan \etal~\cite{fan2023joint} refined GS appearance features together with bidirectional motion fields in a single-stage framework for more efficient RS correction.
However, these methods all rely on supervised learning, which would yield poor generalization on real-world data.
To solve this problem, Qu \etal~\cite{qu2023fast} proposed an efficient geometry-based method, which can correct RS frames according to the optical flows of RS frames.
And Shang \etal~\cite{shang2023self} designed a bidirectional distortion warping module to reconstruct dual reversed RS images, forming self-supervision between input and reconstructed dual reversed RS images.
%
{Then Fan \etal \cite{fan2024self} developed the unified warping model of RS2GS and GS2RS, enabling the complement conversions of RS2GS and GS2RS to be incorporated into a uniform network model.
	%

\subsection{Cycle consistency-based self-supervised learning in low-level vision}
%
The concept of cycle consistency has been utilized in several low-level vision tasks for self-supervised learning.
Zhu~\etal~\cite{zhu2017unpaired} proposed a cycle consistency loss for unpaired image-to-image translation.
%
%
Chen \etal~\cite{chen2018reblur2deblur} enforced the results by fine-tuning existing methods in a self-supervised fashion, where they estimated the per-pixel blur kernel based on optical flows between restored frames, for reconstructing blurry inputs.
Ren \etal~\cite{ren2020neural} utilized two generative networks for respectively modeling the deep priors of clean image and blur kernel to reconstruct blurred image for enforcing cycle consistency.
Liu \etal~\cite{liu2020self} claimed that motion cues obtained from consecutive images yield sufficient information for deblurring task and they re-rendered the blurred images with predicting optical flows for cycle consistency-based self-supervised learning.
Bai \etal~\cite{bai2022self} presented a self-supervised video super-resolution method, which can generate auxiliary paired data from the original low resolution input videos to constrain the network training. 
Shang \etal~\cite{shang2023self} proposed a self-supervised learning method SelfDRSC, which uses a bidirectional distortion warping module to reconstruct dual reversed RS images. Moreover, a self-distillation strategy is proposed to mitigate boundary artifacts in generated GS images.
{La \etal \cite{la2025self} proposed a Cycle-consistent Generative Adversarial Networks (CycleGANs) to complete the Text Style Transfer task in response to the poor robustness caused by the application of cycle-consistency in the latent space and the known issues of long-term text dependency exposed by existing methods relying on recurrent networks.
	%
}
To sum up, in these methods, degraded images can be reconstructed based on the imaging mechanism, and then self-supervised loss can be employed to enforce the cycle consistency between reconstructed and degraded images. 
In this work, we further refine two main components of SelfDRSC. We integrate a joint optimization approach from VFI methods for both optical flows and features to further reduce the parameters of the correction network. We then introduce a VFI-base RS image reconstruction that replaces the bidirectional distortion warping module in SelfDRSC, enhancing the correction effect and avoiding boundary artifacts.

\subsection{Video frame interpolation}
With the resurgence of deep learning, frame rate upconversion has reemerged in the field of computer vision under the name of video frame interpolation.
%
%
%
%
%
%
Huang~\etal~\cite{huang2022rife} and Kong \etal~\cite{kong2022ifrnet} designed an efficient pyramid network and utilized the so-called privileged distillation scheme for real-time arbitrary timestamp frame interpolation.
%
%
Shang \etal~\cite{shang2023joint} considered the motion-induced blur in video sequences and achieved joint video frame interpolation and deblurring.
Recently, advanced network modules and operations have been proposed to further enhance the performance of VFI. For instance, the unified operation in EMA-VFI~\cite{zhang2023extracting} explicitly separates motion and appearance information, and the AMT~\cite{li2023amt} employs bidirectional correlation volumes for all pixel pairs, which are inspired by RAFT~\cite{teed2020raft}.
Zhong \etal~\cite{zhong2023clearer} introduced distance indexing and an iterative reference-based estimation strategy in a plug-and-play manner to enhance the performance of video interpolation.
{All these methods would encounter difficulties when input images are affected by RS distortion, because they cannot correctly synthesize in-between frames without modeling scanline-dependent exposure times. }

\section{Proposed method}\label{sec:method}
In this section, we initially present the problem formulation of self-supervised learning for dual reversed RS correction. Subsequently, we provide a comprehensive introduction to the architecture of the correction network and elaborate on the specifics of the self-supervised learning strategy.

\subsection{Formulation of SelfDRSC++}
Recently, dual reversed RS imaging setting~\cite{zhong2022bringing,shang2023self} has been proposed to address the ambiguity issue in consecutive RS images, where two RS images are captured simultaneously by different scanning patterns, \ie, top-to-bottom ($\bm I_{t2b} \in \mathbb{R}^{H\times W \times 3}$) and bottom-to-top ($\bm I_{b2t} \in \mathbb{R}^{H\times W \times 3}$), as shown in Fig. \ref{fig:dual RS}. 
We first give a formal imaging formation of dual RS images with $H$ rows. 
Without loss of generality, we define the acquisition time $t$ as the midpoint of the whole exposure period, \ie, each RS image is captured from $t_1$ to $t_H$, having $H-1$ readout instants $\tau$, where start scanning time $t_1=t-\tau (H-1)/2$ and end scanning time $t_H=t+\tau (H-1)/2$.
Dual reversed RS images captured at time $t$ can be defined as 
\begin{equation}\label{eq:dual RS}
	\begin{aligned}
		\vspace{-2mm}
		\bm{I}^{(t)}_{t2b}[i] = \bm{I}_g^{(t+\tau(i-(H+1)/2))}[i], \quad
		\bm{I}^{(t)}_{b2t}[i] = \bm{I}_g^{(t+\tau(i-(H+1)/2))}[H-i+1], 
		\vspace{-2mm}
	\end{aligned}    
\end{equation}
where $\bm{I}_g$ is the latent GS image. 
When scanning $i$-th rows for $\bm{I}^{(t)}_{t2b}$ and $\bm{I}^{(t)}_{b2t}$, the image contents are captured from $\bm{I}_g$ at the same instant time $t_i$ but with reversed scanning directions. 
In the following, the superscripts in $\bm{I}_{t2b}^{(t)}$ and $\bm{I}_{b2t}^{(t)}$ are omitted, since $t=(t_1+t_H)/2$.

Under the dual reversed RS imaging setting, RS distortion can be well distinguished, \ie, dual reversed RS images $\bm{I}^{}_{t2b}$ and $\bm{I}^{}_{b2t}$ provide cues for reconstructing GS images between $t_1$ and $t_H$, even for the ambiguous case in the consecutive video setting. 
In the most recent state-of-the-art method IFED \cite{zhong2022bringing}, fully supervised learning is employed for learning DRSC network, where high framerate GS video frames should be collected as ground-truth supervision.  
Albeit obtaining promising performance, high framerate GS frames are not trivial to collect. 
The common solution is to synthesize training datasets where video frame interpolation is usually adopted to increase video framerate \cite{zhong2022bringing}, thus restricting their performance in real-world cases. 

\begin{figure}[!t]\footnotesize
	\centering
	\includegraphics[width=0.8\linewidth]{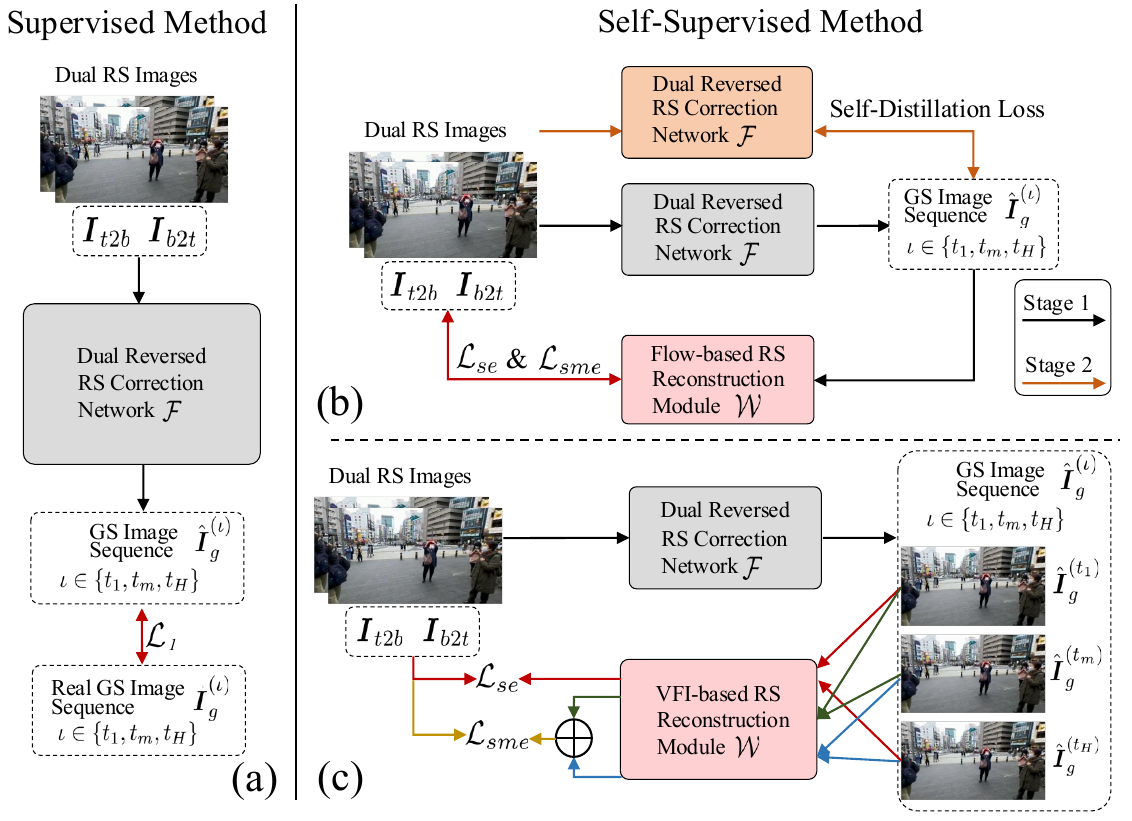}\\
	\vspace{-5mm}
	\caption{
		{
			Comparison of learning frameworks for dual reversed RS correction.
			(a) Supervised methods such as IFED require real GS image sequences as ground truth.
			(b) SelfDRSC performs self-supervised learning with a flow-based RS reconstruction module and a two-stage self-distillation strategy.
			(c) SelfDRSC++ introduces a VFI-based RS reconstruction module guided by distorted time maps, enabling one-stage self-supervised training.
			The predicted GS images $\{\hat I_g^{(t_1)}, \hat I_g^{(t_m)}, \hat I_g^{(t_H)}\}$ are used to reconstruct the input dual RS images and impose the cycle-consistency losses $\mathcal L_{se}$ and $\mathcal L_{sme}$ without real GS supervision.
		}
	}
	\vspace{-3mm}
	\label{fig:selfsup}
\end{figure}
In this work, we propose a novel self-supervised learning method, SelfDRSC++, for rolling shutter correction with dual reversed distortion, where only RS images are required for training DRSC network, without the need for ground-truth high framerate GS images. 
Formally, the optimization of our method is defined as
\begin{equation}
	\vspace{-2mm}
	\underset{\bm\Theta}{ \min } \ \mathcal{L}\left(\left\{\bm I_{t 2 b}, \bm I_{b 2 t}^{}\right\}, \mathcal{W}\left(\mathcal{F}\left(\bm I_{t 2 b}^{}, \bm I_{b 2 t}^{}; \bm \Theta \right)\right)\right),
\end{equation}
which contains two key components: dual reversed RS correction network $\mathcal{F}$ with parameters $\bm{\Theta}$, a VFI-based RS reconstruction module $\mathcal{W}$. 
And then self-supervised learning objective $\mathcal{L}$ can be adopted to enhance the consistency between input and reconstructed RS images. 
{Here, $\mathcal W$ serves as a fixed differentiable surrogate of the GS-to-RS formation process, so the self-supervision is a physically constrained RS--GS--RS cycle rather than a generic image reconstruction loss.}
Our self-supervised framework is depicted in Fig.~\ref{fig:selfsup}{, together with the comparison to supervised training and the previous SelfDRSC framework}.
Different from SelfDRSC~\cite{shang2023self}, we employ a joint optimization of optical flows and corrected features to further reduce parameters of module $\mathcal{F}$. Additionally, we adopt the correlation matching block to assist in refining the estimated optical flows.
For the RS reconstruction module $\mathcal W$, we employ a VFI method based on distorted time maps, {where the distorted time map specifies the row-wise interpolation time induced by RS readout}.
{As analyzed in Sec.~\ref{sec:abl}, this VFI-based reconstruction design provides more faithful RS reconstruction and thus improves the effectiveness of self-supervised training.}

\subsection{Network architecture of DRSC $\mathcal{F}$}
The architecture of $\mathcal{F}$ 
consists of a multi-scale bidirectional RS correction module and a multi-field GS reconstruction module, as shown in Fig.~\ref{fig:framework}. 
%
During training phase, our distortion correction network $\mathcal{F}$ generates three images $\{\hat{\bm{I}}_g^{(t_1)}, \hat{\bm{I}}_g^{(t_m)}, \hat{\bm{I}}_g^{(t_H)}\}$ where $t_m$ is an intermediate scanning time between start time $t_1$ and end time $t_H$. 
During the inference phase, it allows the distortion correction network to generate videos with arbitrary framerate by giving different intermediate scanning times. 
In the following, we take an example to show how intermediate GS image $\hat{\bm{I}}_g^{(t_m)}$ is predicted from dual reversed RS images $\bm{I}_{b2t}$ and $\bm{I}_{t2b}$. 
The start and end GS images $\hat{\bm{I}}_g^{(t_1)}$ and $\hat{\bm{I}}_g^{(t_H)}$ can be obtained in the same way.

\subsubsection{RS correction module}
\begin{figure*}[!t]\footnotesize
	\centering
	\includegraphics[width=\linewidth]{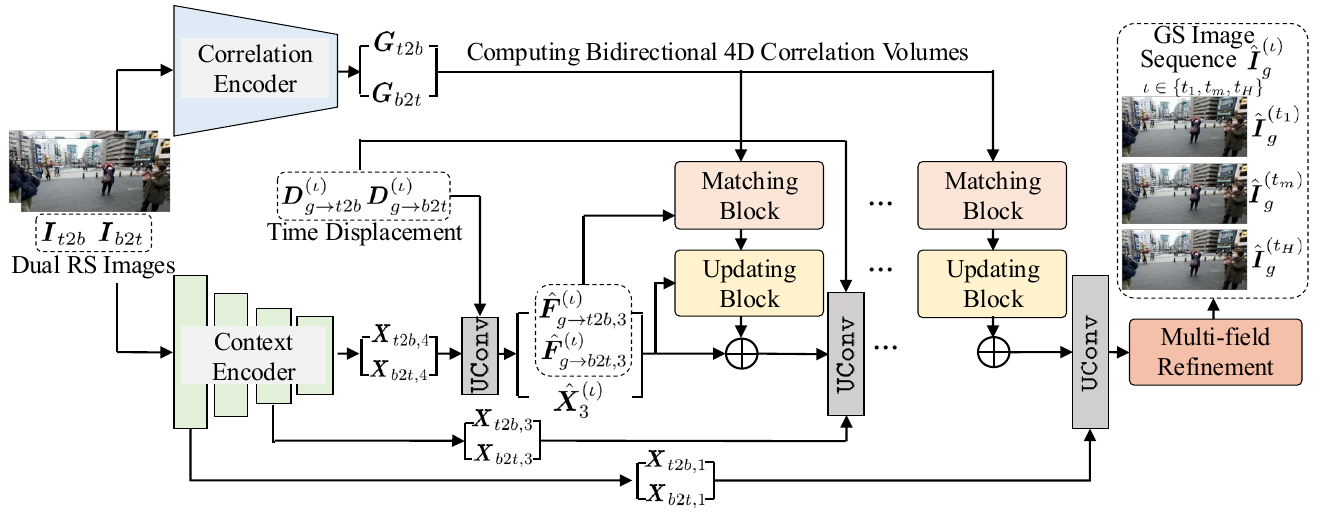}\\
	\vspace{-3mm}
	\caption{
		Architecture of the DRSC network $\mathcal{F}$. The network can correct dual RS images to GS images according to time displacements. It primarily consists of two encoders that extract correlation features $\{\bm{G}_{t2b}, \bm{G}_{b2t}\}$, and content features $\{\bm{X}_{t2b, d}, \bm{X}_{b2t, d}\} (d \in \{4,3,2,1\})$, followed by a decoder with each scale comprising a joint upsampling block ($\mathtt{UConv}$), a matching block, and an updating block. The final output from the last decoder layer is integrated through a multi-field refinement Eq.~(\ref{eq:drsc}) to form the final GS image. {Detailed illustrations of $\mathtt{UConv}$, the Matching Block, and the Updating Block are provided in Fig.~\ref{fig:network_detail}.}
	}
\vspace{-3mm}
	\label{fig:framework}
\end{figure*}
To ensure the interaction between complementary motion and contextual features, we adopt a multi-scale bidirectional RS correction module to update the bidirectional flows and jointly extract intermediate features.
Specifically, our module comprises two encoders and a decoder. The two encoders are designed to encode distinct features, where one encoder is tasked with extracting correlation features to compute bidirectional correlation volumes, while the other is dedicated to capturing content features for predicting corrected features.
We in the following introduce the two encoders and the decoder in detail.

The correlation encoder comprises 4 scales, with each scale consisting of three blocks. Each block within the encoder contains a convolutional layer, followed by an instance normalization layer, and concludes with a ReLU activation function.
We compute the dot-product similarities between the output features $\bm G_{t2b}, \bm G_{b2t} \in \mathbb{R}^{\frac{H}{8}\times \frac{W}{8} \times C}$ for constructing a 4D correlation volume, and this process can be written as
\begin{equation}
	\bm {V}[i, j, k, l]=\sum_h \bm {G}_{t2b}[i, j, h] \cdot \bm{G}_{b2t}[k, l, h], 
\end{equation}
where $\bm{V} \in \mathbb{R}^{\frac{H}{8} \times \frac{W}{8} \times \frac{H}{8} \times \frac{W}{8}}$. {Here, $(i,j)$ and $(k,l)$ denote spatial indices on $\bm{G}_{t2b}$ and $\bm{G}_{b2t}$, respectively, and $h$ denotes the channel index.}
In order to compute the correlation volume at different scales, we perform average pooling on the last two dimensions of $\bm {V}$ to achieve downsampling, with a kernel size of 2 and a stride of 2.
We thus obtain multi-scale correlation volumes $\{\bm {V}_1, \bm {V}_2, \bm {V}_3, \bm {V}_4\}$.
To obtain the bidirectional correlation volume, and to avoid redundant computations, we transpose the matrix $\bm {V}$ to obtain $\bm {V}^{T}$, which serves as the reversed correlation volume.
Also, we can get the multi-scale reversed correlation volumes $\{\bm {V}^T_1, \bm {V}^T_2, \bm {V}^T_3, \bm {V}^T_4\}$.
Correspondingly, our context encoder also generates features at four scales for each RS, denoted as $\{\bm {X}_{t2b,1}, \bm {X}_{t2b,2}, \bm {X}_{t2b,3}, \bm {X}_{t2b,4}\}$ and $\{\bm {X}_{b2t,1}, \bm {X}_{b2t,2}, \bm {X}_{b2t,3}, \bm {X}_{b2t,4}\}$. This enables matching and searching with the correlation volumes in each scale of the subsequent decoder, thereby jointly updating the estimated optical flows and the corrected features.
The design of the context encoder is quite similar to that of the correlation encoder, with the exception that the instance normalization layer has been removed, and the ReLU activation function has been replaced with PReLU.

\begin{figure*}[!t]\footnotesize
	\centering
	\includegraphics[width=0.8\linewidth]{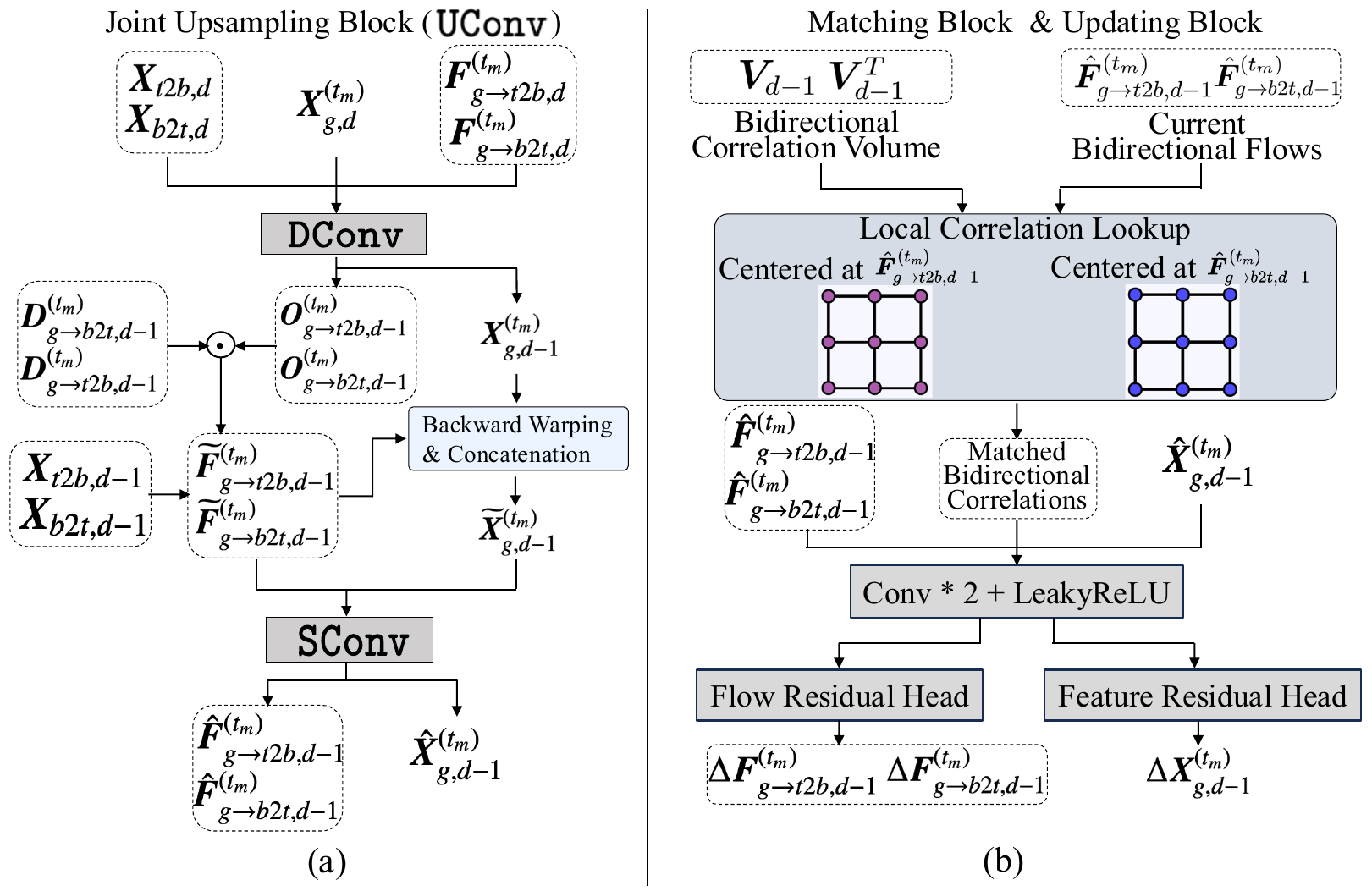}\\
	\vspace{-5mm}
	\caption{
		{{Detailed illustration of the key components.
			(a) The joint upsampling block $\mathtt{UConv}$ consists of $\mathtt{DConv}$ and $\mathtt{SConv}$, jointly updating bidirectional flows and corrected features. 
			(b) The Matching Block queries the bidirectional 4D correlation volumes using the current bidirectional flows, and the Updating Block predicts bidirectional flow residuals and the corrected-feature residual for further refinement.}}
	}
	\vspace{-3mm}
	\label{fig:network_detail}
\end{figure*}
The decoder also contains 4 scales. Except for the last scale, each scale contains a joint upsampling block, a correlation volume matching block, and an updating block.
Given that the decoder aims to jointly optimize the optical flows and the corrected features, it is imperative to provide temporal information. Unlike video frame interpolation tasks, which only require the time instance to interpolate the frame, due to the imaging mechanism of RS, we can obtain the time displacement between the input RS image and the target GS image. 
Formally, the values at $i$-th row in time displacement are given by
\begin{equation}\small
	\begin{aligned}
		\bm D^{\left(t_m\right)}_{g\rightarrow t2b}[i]\!&=\!\frac{i-m}{H-1},\!&i,m \in [1,\!\cdots,\!H], \\
		\bm D^{\left(t_m\right)}_{g\rightarrow b2t}[i]\!&=\!\frac{(H\!-\!i)\!-\!(m\!-\!1)}{H-1},\!&i,m \in [1,\!\cdots,\!H]. 
	\end{aligned}
\end{equation}
This temporal displacement is crucial for the decoder to effectively integrate temporal cues into the optimization process. 
Then we elaborate on each block in detail, taking the $d$-th scale ($d \in \{4,3,2\}$) as an example:
\begin{equation}\label{eq:uconv}
	\begin{aligned}
		\bm F_{g \rightarrow t2b, d-1}^{(t_m)}, \bm F_{g \rightarrow b2t, d-1}^{(t_m)}, \bm X_{g, d-1}^{(t_m)} =
		\mathtt{UConv}(\bm F_{g \rightarrow t2b, d}^{(t_m)}, \bm F_{g \rightarrow b2t, d}^{(t_m)}, \bm X_{g, d}^{(t_m)}, \bm X_{t2b, d}, \bm X_{b2t, d}),
	\end{aligned}
\end{equation}
where $\mathtt{UConv}(\cdot)$ denotes the joint upsampling block. 
{As illustrated in Fig.~\ref{fig:network_detail}(a), $\mathtt{UConv}$ consists of $\mathtt{DConv}$ and $\mathtt{SConv}$, which first predict target-time-aware bidirectional flows and then refine the warped-and-concatenated features.}
Specifically, within the $\mathtt{UConv}(\cdot)$, the inputs -- optical flows, RS features, and corrected RS features -- are initially processed through $\mathtt{DConv}(\cdot)$ to obtain relative motion maps between the latent GS images and the input RS images:
\begin{equation} 
	\begin{aligned}
		\bm O_{g \rightarrow t2b, d-1}^{(t_m)}, \bm O_{g \rightarrow b2t, d-1}^{(t_m)}, {\bm X}_{g, d-1}^{(t_m)} = 
		 \mathtt{DConv}(\bm F_{g \rightarrow t2b, d}^{(t_m)}, \bm F_{g \rightarrow b2t, d}^{(t_m)}, \bm X_{g, d}^{(t_m)}, \bm X_{t2b, d}, \bm X_{b2t, d}),
	\end{aligned}
\end{equation}
where $\mathtt{DConv}(\cdot)$ denotes one ResBlock and one transposed convolution layer.
$\bm O_{g \rightarrow t2b, d-1}^{(t_m)}$ and $\bm O_{g \rightarrow b2t, d-1}^{(t_m)}$ are estimated relative motion maps between latent GS images and the input RS images.
Then the optical flows between input RS images and latent GS images can be obtained by multiplying corresponding time displacement and relative motion map in an entry-by-entry manner, \ie, $\widetilde{\bm{F}}_{g\rightarrow t2b,d-1}^{(t_m)} = \bm{D}_{g\rightarrow t2b,d-1}^{(t_m)} \odot \bm{O}_{g\rightarrow t2b,d-1}^{(t_m)}$ and $\widetilde{\bm{F}}_{g\rightarrow b2t,d-1}^{(t_m)} = \bm{D}_{g\rightarrow b2t,d-1}^{(t_m)}\odot \bm{O}_{g\rightarrow b2t,d-1}^{(t_m)}$.
We employ bilinear interpolation on time displacement at various scales to ensure that its spatial size is aligned with those of relative motion maps.
The corrected features can be obtained from $\bm{X}_{t2b,d-1}$ and $\bm{X}_{b2t,d-1}$ by the backward warping operation with the optical flows. The corrected features are concatenated with other features along the channel dimension dubbed $\widetilde{\bm{X}}^{(t_m)}_{g,d-1}$, combined with the optical flows, and finally fed into a ResBlock $\mathtt{Sconv}$,
\begin{equation} \label{eq:sconv}
	\begin{aligned}
		\hat{\bm F}_{g \rightarrow t2b, d-1}^{(t_m)}, \hat{\bm F}_{g \rightarrow b2t, d-1}^{(t_m)}, \hat{\bm X}_{g, d-1}^{(t_m)} =
		\mathtt{SConv}(\widetilde{\bm{X}}^{(t_m)}_{g,d-1}, \widetilde{\bm F}_{g \rightarrow t2b, d-1}^{(t_m)}, \widetilde{\bm F}_{g \rightarrow b2t, d-1}^{(t_m)}).
	\end{aligned}
\end{equation}
When $d=4$, due to the absence of relative motion maps, we cannot obtain the optical flow. Consequently, the optical flows $\bm F_{g \rightarrow t2b, d}^{(t_m)}$ and $\bm F_{g \rightarrow b2t, d}^{(t_m)}$ are omitted from the input in Eq.~(\ref{eq:uconv}).

Subsequently, we use a correlation volume matching block to search the bidirectional correlation volumes through predicted optical flows, akin to the procedure in RAFT~\cite{teed2020raft}. {As shown in Fig.~\ref{fig:network_detail}(b), the current bidirectional flows serve as the lookup centers for querying the bidirectional 4D correlation volumes within local windows.} The search operation within these windows is conducted across all scales of the bidirectional correlation volume. The matched bidirectional correlations are concatenated into a feature map to further jointly update the optical flows and the corrected features.

Finally, we employ an update block to jointly predict the residuals of the optical flows and the corrected features based on the matched bidirectional correlations. 
In each update block, the bidirectional correlation features and the bidirectional optical flows are first passed through two convolutional layers and LeakyReLU activations in between. Then, they are concatenated with the corrected features and then fed into two additional convolutional layers and LeakyReLU activations in between. 
Finally, the output features are
sent to two separate heads for predicting bilateral flow residuals $\Delta\bm F_{g \rightarrow t2b, d-1}^{(t_m)}$, $\Delta\bm F_{g \rightarrow b2t, d-1}^{(t_m)}$, and the corrected features residual $\Delta\bm X_{g, d-1}^{(t_m)}$. Each head is formed by two convolutional layers and LeakyReLU activations in between.
The predicted optical flows residual and the corrected features residual are added back to the output of Eq.~(\ref{eq:sconv}) to complete the update of optical flows and corrected features by
\begin{equation} 
	\begin{aligned}
		{\bm F}_{g \rightarrow t2b, d-1}^{(t_m)} &= \Delta\bm F_{g \rightarrow t2b, d-1}^{(t_m)} + \hat{\bm F}_{g \rightarrow t2b, d-1}^{(t_m)},  \\
		{\bm F}_{g \rightarrow b2t, d-1}^{(t_m)} &= \Delta\bm F_{g \rightarrow b2t, d-1}^{(t_m)} + \hat{\bm F}_{g \rightarrow b2t, d-1}^{(t_m)}, \\ 
		{\bm X}_{g, d-1}^{(t_m)} &= \Delta\bm X_{g, d-1}^{(t_m)} + \hat{\bm X}_{g, d-1}^{(t_m)}. \\ 
	\end{aligned}
\end{equation}
The updated outputs serve as the input for $\mathtt{UConv}$ in the subsequent scale. 

The last scale ($d=1$) of the decoder only contains a joint upsampling block. 
Distinguished from Eq.~(\ref{eq:uconv}), the current formulation yields outputs that encompass image residuals $\bm R^{(t_m)}$ and occlusion masks $\bm M^{(t_m)}$, diverging from the output of corrected features. 
In addition, we adopt a reconstruction design inspired by AMT~\cite{li2023amt}, which generates a group of optical flows, image residuals, and occlusion masks. These elements are curated to facilitate the forthcoming reconstruction of GS images.

\subsubsection{GS reconstruction module}
The remaining issue is how to fuse warped images for reconstructing latent GS image $\bm{I}_g^{(t_m)}$. 
In flow-based RS correction methods~\cite{zhong2022bringing,shang2023self},  the common solution for generating the final frame is
\begin{equation}\label{eq:drsc}
	\hat{\bm I}_{g}^{(t_m)}\!\!=\!\bm R^{(t_m)}\!+\!\bm M^{(t_m)} \!\odot\! \bm W_{t2b}^{(t_m)}\!+\!(\bm 1\!-\!\bm M^{(t_m)}\!) \odot\! \bm W_{b2t}^{(t_m)},
\end{equation} 
where $\bm W^{(t_m)}_{t2b}=\mathtt{warp}(\bm{I}_{t2b}, \bm{F}_{g\rightarrow t2b}^{(t_m)})$ 
and $\bm W^{(t_m)}_{b2t}= \mathtt{warp}(\bm{I}_{b2t}, \bm{F}_{g\rightarrow b2t}^{(t_m)})$ are obtained by the input RS images through a backward warping operation.
However, predicting a single pair of flow fields overlooks the fact that each location within occluded regions has numerous potential pixels, thereby constraining the solution space of the results~\cite{li2023amt}. 
At this juncture, 
the predicted occlusion masks and image residuals are used to generate multiple candidate GS reconstructions according to Eq.~(\ref{eq:drsc}).
Subsequently, the obtained results are concatenated along the channel dimension and then adaptively merged and refined through two convolutional layers and LeakyReLU activations in between to finalize the output $\hat{\bm{I}}_g^{(t_m)}$.

\subsection{Self-supervised learning for DRSC}
To learn the parameters $\bm \Theta$ of DRSC network $\mathcal{F}$, we need to introduce supervision on $\{\hat{\bm{I}}_g^{(t_1)}, \hat{\bm{I}}_g^{(t_m)}, \hat{\bm{I}}_g^{(t_H)}\}$. 
Instead of collecting ground-truth GS images, we suggest that supervision can be exploited from the input RS images themselves. 
Generally, we introduce a VFI-based module $\mathcal{W}$ to reconstruct dual reversed RS images as shown in Fig.~\ref{fig:RS gen}, and self-supervised loss $\mathcal{L}$ can be adopted to learn the parameters  $\bm \Theta$ without ground-truth GS images. 
{Since the reconstructed RS images must match both top-to-bottom and bottom-to-top inputs under row-wise distorted time maps, this cycle discourages trivial identity-like or globally smoothed solutions.}

\begin{figure*}[!t]\footnotesize
	\centering
	\includegraphics[width=\linewidth]{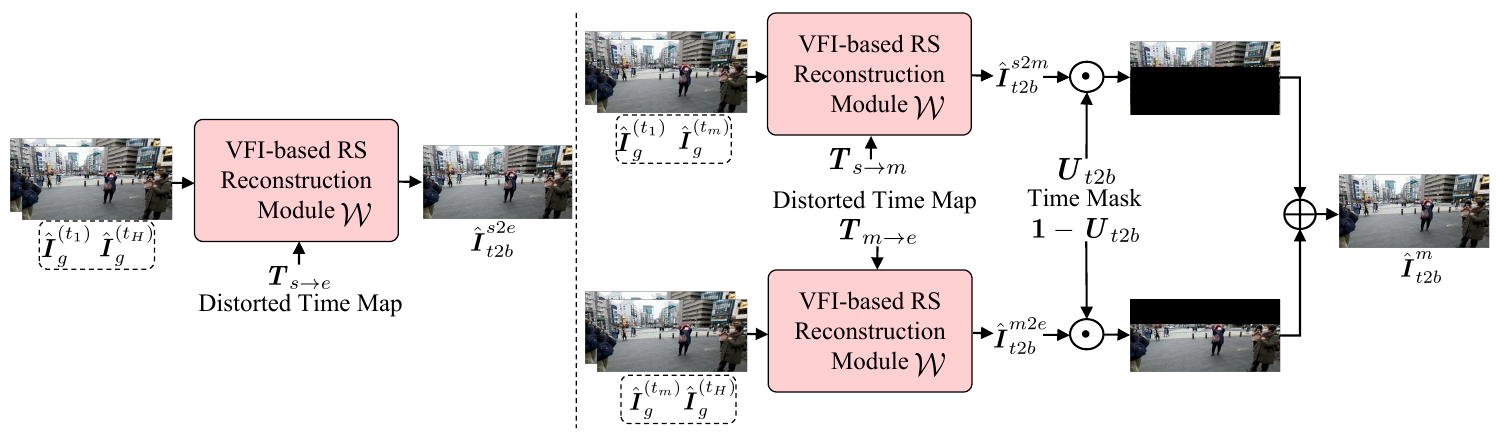}\\
	\vspace{-5mm}
	\caption{
		VFI-based module $\mathcal{W}$ for reconstructing dual reversed RS images. $\mathcal{W}$ is capable of generating the corresponding RS image based on GS inputs and given distorted time maps. 
		{We keep $\mathcal W$ frozen to provide a stable GS-to-RS reconstruction prior; otherwise, jointly optimizing $\mathcal F$ and $\mathcal W$ under only the cycle loss may allow $\mathcal W$ to compensate for imperfect GS predictions without improving their physical correctness.}
	}
\vspace{-3mm}
	\label{fig:RS gen}
\end{figure*}
\subsubsection{Reconstruction of dual reversed RS images}

Based on RS imaging mechanism, generated start and end GS images $\hat{\bm{I}}_g^{(t_1)}$ and $\hat{\bm{I}}_g^{(t_H)}$ can be accordingly exploited to reconstruct dual reversed RS images $\hat{\bm{I}}_{t2b}$ and $\hat{\bm{I}}_{b2t}$. 
Besides start and end GS images, we also provide a way to reconstruct RS images $\hat{\bm{I}}^m_{t2b}$ and $\hat{\bm{I}}^m_{b2t}$ from intermediate GS images $\hat{\bm{I}}_g^{(t_m)}$. 
In the following, we take the top-to-bottom scanning pattern as an example to show the reconstruction of $\hat{\bm{I}}_{t2b}$.

\textit{\textbf{Reconstructing RS images from start and end GS frames $\hat{\bm{I}}_g^{(t_1)}$ and $\hat{\bm{I}}_g^{(t_H)}$.}}
The purpose of $\mathcal{W}$ is to reconstruct RS images $\hat{\bm{I}}_{t2b}$ and $\hat{\bm{I}}_{b2t}$ from $\hat{\bm I}_{g}^{(t_1)}$ and $\hat{\bm I}_{g}^{(t_H)}$. 
Contrary to the approach in~\cite{shang2023self}, we no longer employ optical flow estimation for distorted warping, because the results trained with this method tend to introduce boundary artifacts~\cite{shang2023self}. These artifacts necessitate the use of further self-distillation techniques to mitigate, thereby increasing the complexity and duration of the training process. Consequently, we adopt a more straightforward method to directly obtain the reconstructed RS images.
In \cite{zhong2023clearer}, a video frame interpolation method based on distance indexing is introduced to address the issues arising from the uncertainty of velocity in existing video frame interpolation methods. Interestingly, we regard the generation of RS images as a special type of frame interpolation task, where each row corresponds to a different time instance. We design a distorted time map for representing the interpolation time in each row.
Given the corrected results $\hat{\bm I}_{g}^{(t_1)}$ and $\hat{\bm I}_{g}^{(t_H)}$, we are able to derive a distorted time map $\bm{T}_{s\rightarrow e}$ from the start time to the end time, according to the imaging mechanism of RS:
\begin{equation}\label{eq:time map}
	\bm{T}_{s \rightarrow e}[i]=\frac{(i-1)\cdot\tau}{(H-1)\cdot\tau}=\frac{i-1}{H-1}, i \in [1, \cdots,H].
\end{equation}
And we can also get $\bm T_{e \rightarrow s}=\bm 1 - \bm T_{s \rightarrow e}$. {The distorted time map specifies the temporal coordinate of each scanline, rather than directly approximating motion; the motion synthesis is handled by the frozen VFI-based reconstruction module conditioned on this row-wise time map.}
Due to the extremely short readout time $\tau$ for each row during the actual capture process (\eg, in a 720P video at 30 FPS, the average readout time per row is approximately $\frac{1s}{30\times 720}\approx 46\mu s$). The motion within this extremely short period can be assumed to be linear. We employ the distorted time map in place of the distance indexing used in~\cite{zhong2023clearer} to facilitate the generation of RS images, represented as
\begin{equation}\label{eq:VFI}
	\begin{aligned}
		\hat{\bm{I}}_{t2b}^{s2e} = \mathcal{W}(\hat{\bm I}_{g}^{(t_1)},\hat{\bm I}_{g}^{(t_H)},\bm{T}_{s \rightarrow e}), \quad
		\hat{\bm{I}}_{t2b}^{e2s} = \mathcal{W}(\hat{\bm I}_{g}^{(t_H)},\hat{\bm I}_{g}^{(t_1)},\bm{T}_{e \rightarrow s}).
	\end{aligned}
\end{equation}
Referring to~\cite{zhong2023clearer}, $\mathcal{W}$ can be an existing VFI algorithm, where we directly utilize the pre-trained model provided by~\cite{zhong2023clearer}, without further finetuning on RS correction data.
Following~\cite{shang2023self}, we utilize the distorted time maps as weights to combine the two, yielding the final reconstructed RS image through a weighted summation.
The RS image $\hat{\bm{I}}_{t2b}$ is reconstructed as 
\begin{equation}\label{eq:backwarp}
	\hat{\bm{I}}_{t2b}=\bm{T}_{e\rightarrow s}\odot \hat{\bm{I}}_{t2b}^{s2e} +
	\bm{T}_{s\rightarrow e}\odot \hat{\bm{I}}_{t2b}^{e2s}.
\end{equation}
We can acquire $\hat{\bm{I}}_{b2t}$  in the same manner by vertically flipping the distorted time maps and then applying Eqs.~(\ref{eq:VFI}) and~(\ref{eq:backwarp}) to obtain $\hat{\bm{I}}_{b2t}$.

\textbf{\textit{Reconstructing RS images using intermediate GS frame $\hat{\bm{I}}_g^{(t_m)}$.}}
Aiming to make SelfDRSC++ be able to generate GS frames at the time of arbitrary scanline, we also need to constrain the generated intermediate GS image $\hat{\bm{I}}_g^{(t_m)}$ in a similar way. 
The only distinction is that the generation of RS images is divided into two parts, \ie, one is from $t_1$ to $t_m$ and the other one is from $t_m$ to $t_H$.
According to Eq.~\eqref{eq:time map}, we can obtain distorted time map $\bm T_{s\rightarrow m}$ and $\bm T_{m \rightarrow e}$ as follows
\begin{equation}
	\begin{aligned}
		\bm{T}_{s\rightarrow m}[i]&=\left\{\begin{array}{ll}
			\frac{i-1}{m-1}, &\quad \; i \in[1, \cdots,m] ,\\
			1, &\quad \; i \in[m+1, \cdots,H] ,
		\end{array}\right. 
		\\
		\bm{T}_{m \rightarrow e}[i]&=\left\{\begin{array}{ll}
			0, & i \in[1, \cdots,m], \\
			\frac{i-m-1}{H-m-1}, & i \in[m+1, \cdots,H].
		\end{array}\right.
	\end{aligned}
\end{equation}
Then we can obtain $\hat{\bm I}_{t2b}^{s2m}$ and $\hat{\bm I}_{t2b}^{m2e}$ similar to Eqs.~\eqref{eq:VFI} and~\eqref{eq:backwarp}.
Finally, we can get the reconstructed RS frame $\hat{\bm I}_{t2b}^{m}$ with time mask $\bm U_{t2b}$:
\begin{equation}\label{eq:aggre}
	\hat{\bm I}_{t2b}^{m}=\bm U_{t2b}\odot \hat{\bm I}_{t2b}^{s2m}+(\bm 1-\bm U_{t2b})\odot \hat{\bm I}_{t2b}^{m2e},
\end{equation}
where 
\begin{equation}
	\bm U_{t2b}[i]=\left\{\begin{array}{ll}
		0, & \text{if  } i\textgreater m \\
		1, & \text{else}
	\end{array}\right.  
\end{equation}
Similarly, by vertically flipping the distorted time maps and the time mask, and then according to Eqs.~(\ref{eq:VFI}),~(\ref{eq:backwarp}) and~(\ref{eq:aggre}), we can obtain $\hat{\bm I}_{b2t}^{m}$.

\subsubsection{Self-supervised losses}
We have reconstructed two sets of dual reversed RS images, \ie, $\hat{\bm{I}}_{t2b}$, $\hat{\bm{I}}_{b2t}$ and $\hat{\bm{I}}^m_{t2b}$, $\hat{\bm{I}}^m_{b2t}$, and thus a loss function $\ell$ can be imposed on their corresponding input RS images, where $\ell$ is the combination of Charbonnier loss~\cite{lai2018fast} and perceptual loss~\cite{johnson2016perceptual} in our experiments, and the hyper-parameters for balancing them are set as 1 and 0.1. 
The self-supervised loss function is defined as
\begin{equation}
	\mathcal{L}_{self}=\mathcal{L}_{se}+\mathcal{L}_{sme},
\end{equation}
with 
%
\begin{equation}\label{eq:loss}
	\begin{aligned}
		\mathcal{L}_{se}=\ell(\hat{\bm{I}}_{t2b}, \bm{I}_{t2b})+\ell(\hat{\bm{I}}_{b2t}, \bm{I}_{b2t}^{}), \quad
		\mathcal{L}_{sme}=\ell(\hat{\bm{I}}_{t2b}^{m}, \bm{I}_{t2b}^{})+\ell(\hat{\bm{I}}_{b2t}^{m}, \bm{I}_{b2t}^{}).
	\end{aligned}	
\end{equation}

Unlike~\cite{shang2023self}, our improved model $\mathcal{W}$ directly employs a video frame interpolation method, which does not introduce boundary artifacts as in the bidirectional distortion warping method based on optical flow. Consequently, we no longer require the self-distillation loss in~\cite{shang2023self}, simplifying the two-stage training strategy from~\cite{shang2023self} into a one-stage training strategy.

\section{Experiments}\label{sec:experiments}

In this section, our SelfDRSC++ is evaluated on synthetic and real-world RS images, and animated video results can be found from the link \href{https://1drv.ms/f/c/41432672867a3bd6/EqUoQLKE7-VMtu09mb-CVoEBOZ64d2LCiHjEazZPKthNtQ?e=572nCd}{\textbf{\emph{Video Results}}}.
Furthermore, we conducted a more detailed analysis of modules in SelfDRSC++, including the impact of employing different VFI methods within module $\mathcal{W}$, as well as the effects of VFI methods training with different variants. Additionally, we performed an ablation analysis on the structure within module $\mathcal{F}$.

\subsection{Datasets}

\subsubsection{Synthetic dataset}\label{sec:data}

The synthetic dataset RS-GOPRO from IFED \cite{zhong2022bringing} is used for quantitatively evaluating the competing methods. 
The dataset is randomly split into 50, 13, and 13 sequences as training, validation, and testing sets, respectively.
Furthermore, 9 GS images serve as the ground-truth for both training and testing. This implies that within this dataset, each pair of dual reversed RS images corresponds to 9 ground-truth GS images.
We note that the ground-truth GS images may not be truly collected by a GOPRO camera. 
As discussed in \cite{shang2023self}, a VFI method \cite{huang2022rife} is adopted to generate high framerate GS images, and the interpolated ground-truth GS images may be over-smoothed. 
Therefore, {PSNR and SSIM should be interpreted with caution on RS-GOPRO, since they can favor outputs closer to the interpolated references, while LPIPS, and visual comparisons provide complementary evidence for perceptual quality and temporal consistency.}

\subsubsection{Real-world testing set} 

For testing in real-world, \cite{zhong2022bringing} built a dual reversed RS image acquisition system in Fig.~\ref{fig:dual RS}, which consists of a beam-splitter and two RS cameras with dual reversed scanning patterns. 
Each sample includes two RS distorted images with reversed distortion but without ground-truth. 
{We also captured some real data with different camera parameters (such as different readout $\tau=30 \mu s, 60 \mu s, 90 \mu s, 120 \mu s, 150 \mu s$) in the same way to evaluate the generalization ability of the method in real cases. Each readout setting contains 10 segments of real captured video sequences. }
\begin{figure*}[!t]\footnotesize
	\centering
	\includegraphics[width=0.85\linewidth]{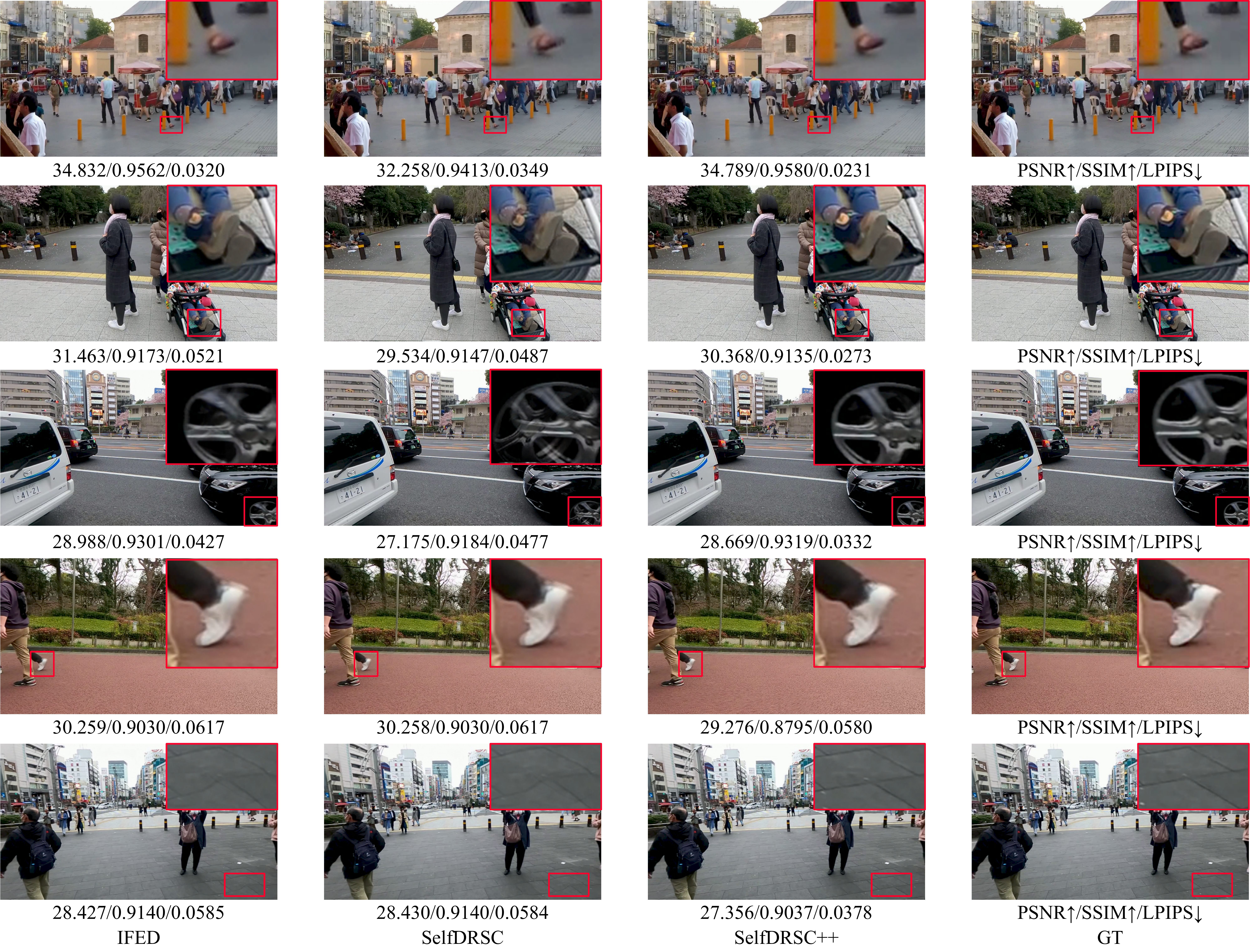}\\
	\vspace{-5mm}
	\caption{Visual comparison on RS-GOPRO. Our SelfDRSC++ yields less ghosting while maintaining fine details. 
	{Since the 9-frame GS references in RS-GOPRO are generated by VFI and may be over-smoothed, PSNR/SSIM can favor results closer to the interpolated references. LPIPS and visual comparisons therefore provide complementary evidence for perceptual quality.} }
	\vspace{-3mm}
	\label{fig:compare}
\end{figure*}
\begin{figure*}[!t]\footnotesize
	\centering
	\includegraphics[width=0.95\linewidth]{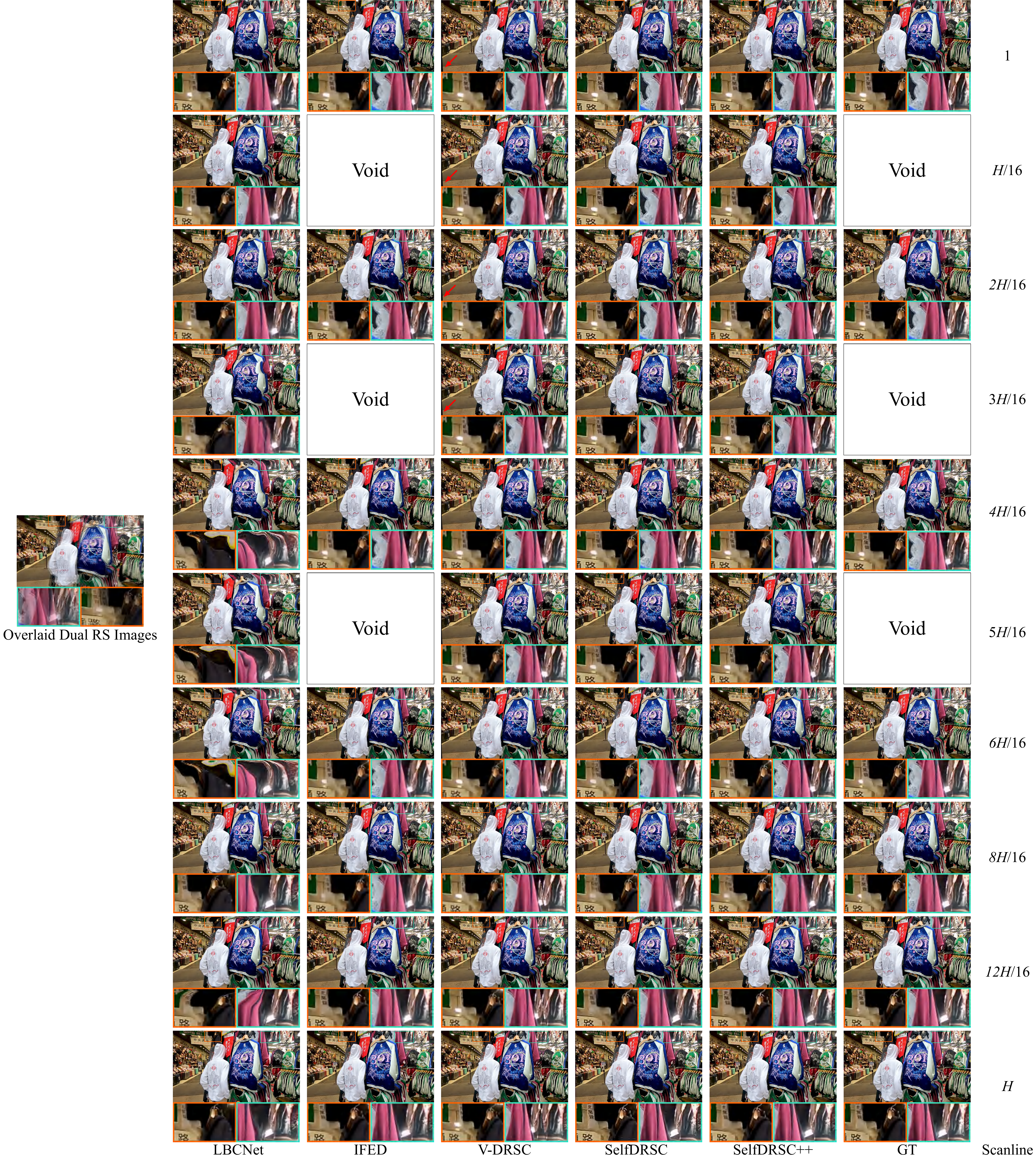}\\
	\vspace{-5mm}
	\caption{Visual results on RS-GOPRO.
		IFED \cite{zhong2022bringing} can only generate a GS video with 9 frames since 9 ground-truth GS images from RS-GOPRO are used to train DRSC network in a supervised manner. 
		Our SelfDRSC++ is able to generate GS videos with a higher framerate. 
		In this case, 17 GS frames are generated by SelfDRSC++. 
		Notably, V-DRSC exhibits black boundaries due to the utilization of backward warping, as indicated by the red arrows in the first four rows. 
		It will be better viewed by zooming in. }
	\vspace{-3mm}
	\label{fig:syn}
\end{figure*}

\subsection{Implementation details} \label{sec: ID}
Our SelfDRSC++ is trained in an end-to-end manner and only the parameters of $\mathcal{F}$ are updated. 
Within $\mathcal{F}$ module, the window size in the matching block is set as 3, which is the same as~\cite{li2023amt}.
The training is done in 150K iterations. 
The optimization is implemented using AdamW~\cite{loshchilov2017decoupled} optimizer ($\beta_1$=0.9, $\beta_2$=0.999) and the initial learning rate is $2 \times 10^{-4}$, which is gradually reduced to $5 \times 10^{-5}$ with the cosine annealing~\cite{loshchilov2016sgdr}.
Patch size is $256\times 256$, and batch size is 16 during the training process. 
The intermediate time $t_m$ is randomly sampled between start time $t_1$ and end time $t_H$ with sampling interval $(t_H-t_1)/8$. 
During testing, we can specify an arbitrary time instance $t_m$ for the correction.

\subsection{Comparison with state-of-the-art methods}
To keep consistent quantitative evaluation with IFED \cite{zhong2022bringing}, the output of competing methods should have 9 GS images. 
Thus, our SelfDRSC++ is compared with methods from three categories. 
(i) The first category contains cascade methods, where RS correction method DUN~\cite{liu2020deep} for generating one GS image and a VFI model RIFE \cite{huang2022rife} is adopted for interpolating GS images.
%
Both of them are re-trained on the RS-GOPRO dataset for a fair comparison. 
In Tables~\ref{tabel:syn} and~\ref{tabel:synval}, both cascade orders are considered, \ie, DUN+RIFE and RIFE+DUN. 
(ii) The second category contains RS correction methods with multiple output images. 
There are only two works by adopting the dual reversed RS correction setting, \ie, IFED \cite{zhong2022bringing} and \cite{albl2020two}. 
Since the source code or experiment results of \cite{albl2020two} are not publicly available, it is not included in comparison. 
Besides, we take RSSR~\cite{Fan_2021_ICCV}, CVR~\cite{Fan_2022_CVPR}, JAM~\cite{fan2023joint}, LBCNet~\cite{fan2024learning} and Uni-SoftSplat~\cite{fan2024unified} into comparison, which are developed for correcting RS distortion from consecutive two RS images with only top-to-bottom scanning. 
For a fair comparison, they are re-trained based on the RS-GOPRO dataset. 
(iii) In addition, we also compare with five unsupervised methods, V-DRSC~\cite{qu2023fast}, SelfDRSC~\cite{shang2023self}, SelfRSSplat~\cite{fan2024self}, {SelfUnroll \cite{lin2025self}, and Self-EvRSVFI \cite{lu2025self}}. 
As for quantitative IQA metrics, PSNR, SSIM~\cite{wang2004image}, and LPIPS~\cite{zhang2018unreasonable} are employed to evaluate the competing methods. 
\begin{table*}[!htb]\footnotesize  
	\centering
	\caption{{Quantitative comparison on the test set of RS-GOPRO \cite{zhong2022bringing}, where the metrics are computed based on generating 9 GS images for a testing case. For event-based self-supervised methods, we use ESIM \cite{rebecq2018esim} to synthesize corresponding events to ensure fair comparisons on the same dataset.}}\label{tabel:syn}
	\resizebox{\linewidth}{!}{
		\begin{tabular}{c|c|c|c|c}  
			
			\hline
			
			\hline
			
			& Method &  Inference time (\textit{s})  & Parameters &   PSNR$\uparrow$ / SSIM$\uparrow$ / LPIPS$\downarrow$  \\  
			\hline
			\multirow{8}{*}{{Supervised}} & DUN~\cite{liu2020deep}+RIFE~\cite{huang2022rife} & 0.638  &   14.62\text{M}  &  23.597 / 0.7653 / 0.1670 \\  
			& RIFE~\cite{huang2022rife}+DUN~\cite{liu2020deep} & 4.078  & 14.62\text{M}  &  20.012 / 0.6520 / 0.1781  \\ 
			&  RSSR~\cite{Fan_2021_ICCV} & 0.976   & 26.03\text{M}  &  22.729 / 0.7283 / 0.1026  \\   
			&  CVR~\cite{Fan_2022_CVPR} & 2.088  & 42.70\text{M}  &   24.816 / 0.7804 / 0.0738  \\   
			&  IFED~\cite{zhong2022bringing} &  0.177& 29.86\text{M}  &   \textbf{30.681} / \textbf{0.9121} / \underline{0.0453}   \\   
			&  JAM~\cite{fan2023joint} &  0.257  &  4.73\text{M}  &    24.905 / 0.7891 / 0.0785   \\ 
			& Uni-SoftSplat~\cite{fan2024unified}  & 1.893    &  7.44\text{M}    &  24.911 / 0.7890 / 0.0790   \\
			&  LBCNet~\cite{fan2024learning} &  0.233  &  10.70\text{M}  &    25.193 / 0.7954 / 0.0762   \\ 
			\hline
			
			\multirow{6}{*}{{Unsupervised}}  &  V-DRSC~\cite{qu2023fast} &  \textbf{0.073}  & 5.26\text{M}  &   25.868 / 0.8338 / 0.0745  \\   
			&SelfDRSC~\cite{shang2023self} & 0.182 & 28.75\text{M} &   28.704 / 0.8886 / 0.0546   \\  
			
			& SelfRSSplat~\cite{fan2024self}  &  1.893  &   7.44\text{M}   &  24.872 / 0.7799 / 0.0782  \\
			
			&SelfUnroll \cite{lin2025self}  &  {0.331}  & {33.00}\text{M}  &   25.911 / 0.8342 / 0.0744 \\
			&Self-EvRSVFI \cite{lu2025self}  &  {0.312}  & {22.00}\text{M}  &   25.921 / 0.8348 / 0.0742 \\
			
			&SelfDRSC++ & 0.156 & \textbf{2.99}\text{M} &    \underline{29.584} / \underline{0.8989} / \textbf{0.0341}    \\   
			\hline
			
			\hline
		\end{tabular}
	}
\vspace{-3mm}
\end{table*}

\subsubsection{Results on synthetic dataset}
Tables \ref{tabel:syn} and \ref{tabel:synval} report quantitative comparison, where the IQA metrics are computed based on 9 GS images for each testing case. 
{We have included a comparison with event-based self-supervised methods (SelfUnroll \cite{lin2025self} and Self-EvRSVFI \cite{lu2025self}). We synthesized corresponding events for the RS-GOPRO dataset using ESIM \cite{rebecq2018esim} to enable fair training and comparison on the same dataset. It can be observed that even with the aid of event data, these methods still do not perform as well as those based on the dual reversed setting. }
Our SelfDRSC++, without any ground-truth high framerate GS images during training, achieves better IQA metrics than the other methods except IFED. 
Although our method exhibits lower PSNR and SSIM compared to IFED, it performs better in terms of LPIPS metric and visual quality in Figs.~\ref{fig:compare} and \ref{fig:syn}. 
As depicted in Fig. \ref{fig:compare}, {PSNR and SSIM should be interpreted with caution, because the interpolated GS references may favor outputs that are closer to the smoothed references.}
In contrast, LPIPS demonstrates a stronger correlation with visual perception. 
However, considering the presence of interpolated frames introduced by the VFI method in 9 ground-truth GS images as discussed in \cite{shang2023self}, these IQA metrics may not provide reliable indications of correction performance. Therefore, we recommend relying on visual results, particularly animated videos, to assess correction textures and temporal consistency.
\begin{table}[!tb]\footnotesize  
	\centering
	\caption{{Quantitative comparison on the validation set of RS-GOPRO \cite{zhong2022bringing}, where the metrics are computed based on 9 GS images for a testing case. For event-based self-supervised methods, we use ESIM \cite{rebecq2018esim} to synthesize corresponding events to ensure fair comparisons on the same dataset.}}\label{tabel:synval}
	\begin{tabular}{c|c|c}  
		\hline
		
		\hline
		& Method &   PSNR$\uparrow$ / SSIM$\uparrow$ / LPIPS$\downarrow$  \\  
		\hline
		\multirow{8}{*}{{Supervised}} & DUN~\cite{liu2020deep}+RIFE~\cite{huang2022rife} &  22.313 / 0.7335 / 0.1907 \\  
		& RIFE~\cite{huang2022rife}+DUN~\cite{liu2020deep} & 19.898 / 0.6481 / 0.2073  \\  
		&  RSSR~\cite{Fan_2021_ICCV} &  22.287 / 0.7076 / 0.1052  \\   
		&  CVR~\cite{Fan_2022_CVPR} &   24.286 / 0.7649 / 0.0762  \\   
		&  IFED~\cite{zhong2022bringing} &    \textbf{30.106} / \textbf{0.9046} / \underline{0.0460}   \\   
		&  JAM~\cite{fan2023joint} &      24.294 / 0.7674 / 0.0749   \\ 
		& Uni-SoftSplat~\cite{fan2024unified}   &  24.295 / 0.7672 / 0.0772   \\
		&  LBCNet~\cite{fan2024learning} &      24.433 / 0.7692 / 0.0741   \\ 
		\hline
		
		\multirow{6}{*}{{Unsupervised}}  &  V-DRSC~\cite{qu2023fast} &  24.470 / 0.7763 / 0.0850  \\   
		&SelfDRSC~\cite{shang2023self} &  27.818 / 0.8758 / 0.0581   \\  
		& SelfRSSplat~\cite{fan2024self}    &  24.034 / 0.7660 / 0.0821  \\
		&SelfUnroll \cite{lin2025self}  &   25.002 / 0.7911 / 0.0759 \\
		&Self-EvRSVFI \cite{lu2025self}  &   25.013 / 0.7912 / 0.0758 \\
		&SelfDRSC++ &   \underline{28.888} / \underline{0.8898} / \textbf{0.0354}    \\   
		\hline
		
		\hline
	\end{tabular}
	\vspace{-1mm}
\end{table}
\begin{figure*}[!t]\footnotesize
	\centering
	\includegraphics[width=\linewidth]{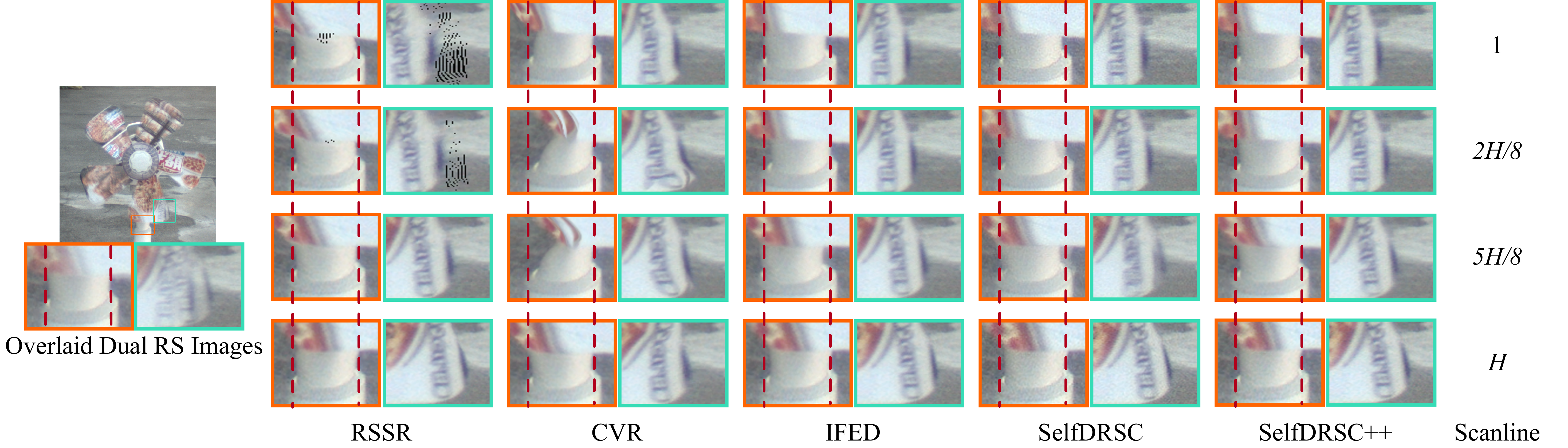} 
	\vspace{-8mm}
	\caption{Comparison of corrected results for moving and stationary objects on real-world RS data.
		Indicated by red dashed lines, stationary objects corrected by other methods exhibit deformation. As green boxes contain objects in rapid motion, other methods demonstrate abnormal distortion or ghosting effects. In contrast, our SelfDRSC++ outperforms other methods for both stationary and moving objects.
		Animated videos are available at \href{https://1drv.ms/f/c/41432672867a3bd6/EqUoQLKE7-VMtu09mb-CVoEBOZ64d2LCiHjEazZPKthNtQ?e=572nCd}{\textbf{\emph{Video Results}}}. 
	}
\vspace{-3mm}
	\label{fig:real}
\end{figure*}
\begin{table}[!htb]\footnotesize  
	\setlength{\tabcolsep}{5pt}
	\centering
	\setlength{\abovecaptionskip}{0pt} 
	\setlength{\belowcaptionskip}{0pt}
	\caption{{Quantitative comparison on real-world RS data using NR-IQA metrics and the motion-compensated temporal consistency metric $E_{\mathrm{temp}}$.}	}\label{tabel:real}
	\begin{tabular}{c|c|c|c}
		\hline
		
		\hline	
		& Method &  NIQE$\downarrow$ / NRQM$\uparrow$ / PI$\downarrow$  &   $E_{\mathrm{temp}}$$\downarrow$   \\
		
		\hline
		\multirow{8}{*}{{Supervised}} & DUN~\cite{liu2020deep}+RIFE~\cite{huang2022rife} &  5.012 / 5.719 / 4.895 & 0.3098 \\  
		& RIFE~\cite{huang2022rife}+DUN~\cite{liu2020deep} & 5.110 / 5.681 / 4.732  & 0.3276  \\  
		&  RSSR~\cite{Fan_2021_ICCV} &  {4.495} / {6.958} / {3.794}  & 0.1566\\   
		&  CVR~\cite{Fan_2022_CVPR} &   4.950 / 5.933 / 4.599 &  0.1122\\   
		&  IFED~\cite{zhong2022bringing} &    4.873 / 5.953 / 4.611  & \underline{0.0758}\\   
		&  JAM~\cite{fan2023joint} &      4.892 / 5.940 / 4.592   & 0.1106 \\ 
		& Uni-SoftSplat~\cite{fan2024unified}   &  4.888 / 5.912 / 4.571  & 0.1131 \\
		&  LBCNet~\cite{fan2024learning} &      4.893 / 5.961 / 4.546  & 0.1086 \\ 
		\hline
		
		\multirow{4}{*}{{Unsupervised}}  &  V-DRSC~\cite{qu2023fast} &  4.400 / 7.661 / 3.650  & 0.1139 \\   
		&SelfDRSC~\cite{shang2023self} &  \underline{4.391} / \underline{7.683} / \underline{3.646} &  0.0783 \\  
		& SelfRSSplat~\cite{fan2024self}    &  4.489 / 6.912 / 3.774 & 0.1296 \\
		
		&SelfDRSC++ &   \textbf{4.262} / \textbf{7.796} / \textbf{3.477}  &  \textbf{0.0472} \\   
		\hline

		\hline
		
		\hline
	\end{tabular}
	\vspace{-3mm}
\end{table}

As for the visual quality, reconstructed GS images by competing methods are presented in Fig. \ref{fig:syn}. 
IFED \cite{zhong2022bringing} can only generate a GS video with 9 frames. 
Our SelfDRSC++ is able to generate GS videos with a higher framerate. 
In this case, 17 GS frames are generated by SelfDRSC++ and are better corrected with finer textures. 
Compared to V-DRSC~\cite{qu2023fast}, our results do not exhibit the boundary content loss caused by backward warping operations. In contrast to SelfDRSC~\cite{shang2023self}, our corrected results are more accurate, as can be observed in the green boxes at $t_m = 6H/16$ and $8H/16$. In summary, our SelfDRSC++ ensures the correction capability of SelfDRSC while further addressing the inaccuracy of SelfDRSC results near $t_m = H/2$.
{Moreover, our method has advantages in both the number of parameters and runtime.}


\subsubsection{Results on real-world data}
In Fig.~\ref{fig:real}, we provide an example of real-world dual reversed RS images. 
By examining the overlaid dual RS images, it is evident that the fan pole within the orange box remains stationary, while the fan blades within the green box are rotating at high speed. However, existing correction methods struggle to accurately discern whether objects in the images are distorted. 
All methods except our SelfDRSC++ have resulted in displacement or distortion, as compared to the orange box in Fig.~\ref{fig:real}. For objects in rapid motion, such as those in the green box, other methods exhibit abnormal distortion or ghosting at object boundaries. Our SelfDRSC++ ensures that stationary objects are not subject to deformation while effectively correcting the distortion for moving objects.
The link \href{https://1drv.ms/f/c/41432672867a3bd6/EqUoQLKE7-VMtu09mb-CVoEBOZ64d2LCiHjEazZPKthNtQ?e=572nCd}{\textbf{\emph{Video Results}}} provides more video results for further comparisons of visual quality and temporal consistency.   

{On the other hand, we use no-reference image quality assessment (NR-IQA) metrics (i.e., NIQE~\cite{mittal2012making}, NRQM~\cite{ma2017learning} and PI~\cite{blau20182018}) to quantitatively compare the results of our method on real data. {We further report a motion-compensated temporal consistency metric $E_{\mathrm{temp}}$~\cite{ruder2016artistic} to evaluate the consistency of generated GS sequences without GS references.} Table~\ref{tabel:real} shows that our method consistently achieves better {NR-IQA scores and the lowest $E_{\mathrm{temp}}$}, indicating superior perceived quality.}

{
	\subsubsection{Generalization ability on different readout $\tau$}
	One key factor affecting RS imaging across different camera parameters is the per-row readout time $\tau$. 
	We also captured some real data with different camera parameters (such as different $\tau$) in the same way to evaluate the generalization ability of the method in real cases. Animated video results can be found from the link \href{https://1drv.ms/f/c/41432672867a3bd6/EqUoQLKE7-VMtu09mb-CVoEBOZ64d2LCiHjEazZPKthNtQ?e=572nCd}{\textbf{\emph{Video Results}}}.
	Moreover, to better verify the method's generalization across different readout settings, we created and tested data with diverse readout configurations using synthetic datasets. The results are shown in Fig. \ref{fig:readout2}. The learned IFED suffers from texture loss in corrected results (see hair and characters in Fig. \ref{fig:readout2}), while our method demonstrates good generalization with finer textures. Additionally, our method shows stronger correction ability compared to the previous version, such as being closer to the GT when the readout is 195 $\mu s$. 
}
\begin{figure*}[!t]\footnotesize
	\centering
	\includegraphics[width=0.75\linewidth]{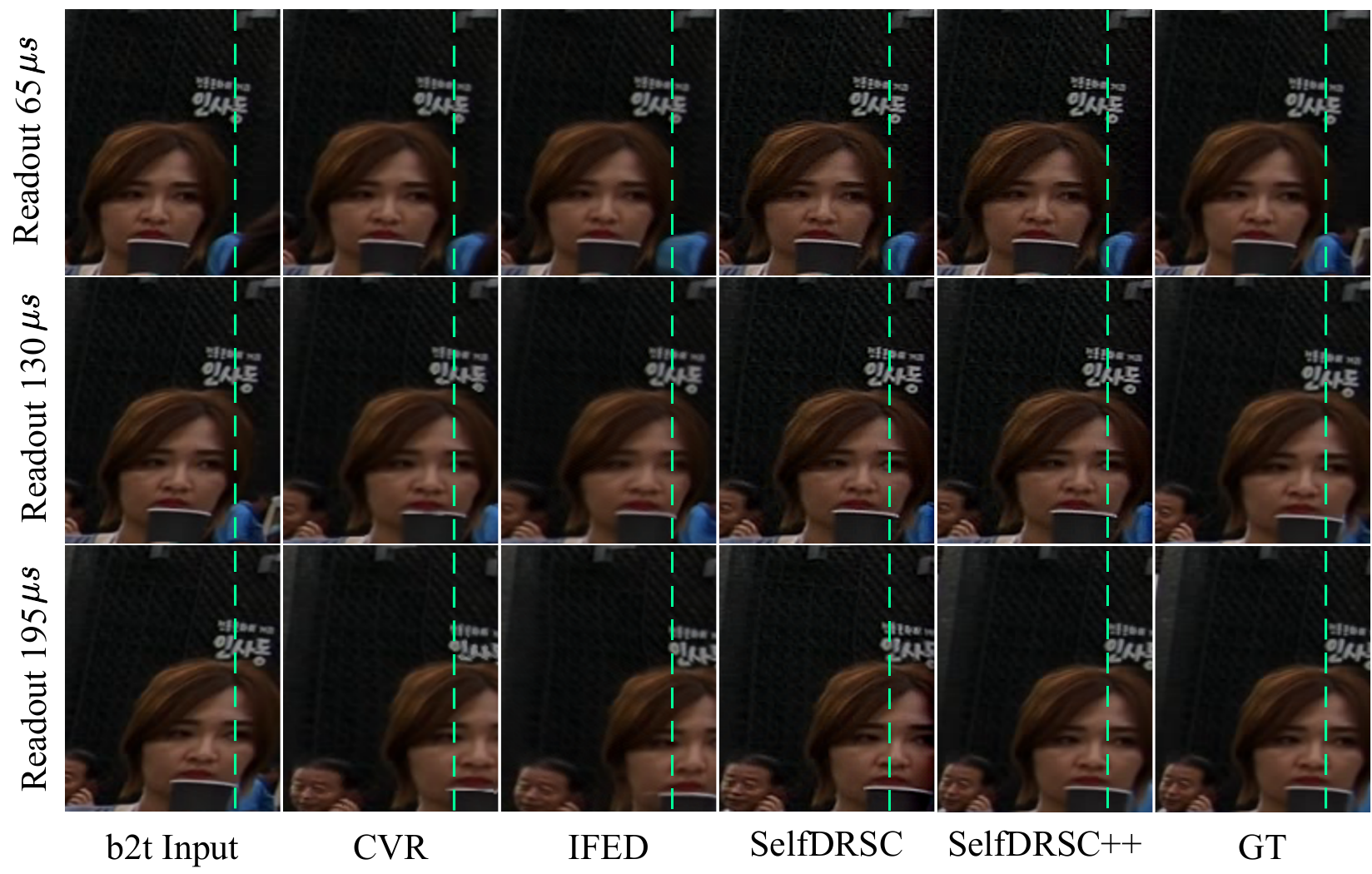}\\
	\vspace{-5mm}
	\caption{{Generalization ability on different readout $\tau$. All methods are trained on fixed readout $\tau=87 \mu s$, while our method can generalize better than others under different readouts.}}
	\vspace{-5mm}
	\label{fig:readout2}
\end{figure*}

\subsection{Ablation study}\label{sec:abl}
For the RS reconstruction module $\mathcal{W}$, we demonstrate the effectiveness of different VFI methods and various variants for VFI methods. 
For the RS correction module $\mathcal{F}$, we also examined the effectiveness of the correlation matching module, the number of fields, {and the window size $w$ of the bidirectional correlation matching block} in the GS reconstruction module.
We implement the ablation study on these elements. All the IQA metrics are computed based on generating 9 GS images.

For each VFI method $\mathcal{W}$, we can combine the plug-and-play nature of distance indexing and iterative reference-based estimation strategies~\cite{zhong2023clearer} to form various variants for VFI methods. 
Specifically, we denote the variant trained with the time indexing as [T], the variant trained with distance indexing as [D], the variant trained with iterative reference-based estimation as [T, R], and the variant that combines both distance indexing and iterative reference-based estimation as [D, R].
All these trained models are borrowed from~\cite{zhong2023clearer} without finetuning on the RS dataset. We replaced the time and distance indexing with our proposed distorted time map as $\mathcal{W}$ to reconstruct RS for self-supervised learning.
We take RIFE~\cite{huang2022rife} as an example. 
As shown in Table~\ref{tabel:abla}, [D, R] RIFE gets the best performance, since it combines both distance indexing and iterative reference-based estimation to solve the ambiguity of speed and direction.
In Fig.~\ref{fig:varient}, we present a visual comparison of RS reconstruction results for various variants of RIFE. Our observations reveal that the reconstructions by [D, R] RIFE more closely resemble the original RS images, thus indicating that correction module $\mathcal{F}$, trained with [D, R] RIFE (as $\mathcal{W}$), yields the most effective performance.
\begin{table}[!t]\footnotesize  
	\centering
	\caption{Ablation results of RS reconstruction module $\mathcal{W}$ on the test set of RS-GOPRO. {[T], [D], and [R] denote the time index, distance index, and iterative reference-based estimation strategy in VFI, respectively.}}\label{tabel:abla}
	\begin{tabular}{cc|c}
		\hline
		
		\hline
		Method &  Strategy &  PSNR$\uparrow$ / SSIM$\uparrow$ / LPIPS$\downarrow$  \\
		
		\hline
		\multirow{4}{*}{{RIFE}}  &   [T]  &  27.761 / 0.8746 / 0.0508  \\
		&  [D]  &  28.870 / 0.8755 / 0.0358 \\
		&  [T, R]  &  26.534 / 0.8456 / 0.0510  \\
		&   [D, R]  &  \textbf{29.584} / \textbf{0.8989} / \textbf{0.0341}   \\
		
		\hline
		
		  EMA-VFI  &  \multirow{4}{*}{{[D, R]}}  & 25.296 / 0.8423 / 0.1016  \\
		
		IFRNet &  &     25.160 / 0.7554 / 0.0983  \\
		AMT  &   &    24.308 / 0.7011 / 0.0753    \\
		RIFE   &   &    \textbf{29.584} / \textbf{0.8989} / \textbf{0.0341}  \\
		\hline
		
		\hline
	\end{tabular}
\vspace{-3mm}
\end{table}
\begin{table}[!t]\footnotesize  
	\centering
	\caption{{Ablation study of RS correction module $\mathcal{F}$ in SelfDRSC++. {`num' denotes the number of reconstruction fields in multi-field refinement, and $w$ denotes the matching window size.}}}\label{tabel:abla2}
		\setlength{\tabcolsep}{4pt}
		\begin{tabular}{cc|c}
			\hline
			
			\hline
			\multicolumn{2}{l|}{Variant} &  PSNR$\uparrow$ / SSIM$\uparrow$ / LPIPS$\downarrow$  \\
			
			\hline
			\multicolumn{2}{l|}{w/o matching \& updating block}   &  28.455 / 0.8814 / 0.0386  \\
			\hline
			\multicolumn{2}{l|}{SelfDRSC++ (num=1)}    &  29.233 / 0.8939 / 0.0375 \\
			\multicolumn{2}{l|}{SelfDRSC++ (num=3) (Ours)}    & {\textbf{29.584} / \textbf{0.8989} / \textbf{0.0341}}   \\  
			\multicolumn{2}{l|}{SelfDRSC++ (num=5)}   &  {29.379} / {0.8972} / {0.0352}   \\
			
			\multicolumn{2}{l|}{SelfDRSC++ (num=7)}   &  {29.395} / {0.8966} / {0.0349}   \\
			\hline
			\multicolumn{2}{l|}{SelfDRSC++ ($w=3$) (Ours)}    &  {29.584} / {0.8989} / {0.0341}   \\
			\multicolumn{2}{l|}{SelfDRSC++ ($w=5$)}   &  {29.597} / {0.8991} / \textbf{0.0340}   \\
			
			\multicolumn{2}{l|}{SelfDRSC++ ($w=7$)}   &  \textbf{29.600} / \textbf{0.8992} / \textbf{0.0340}   \\
			\hline
			
			\hline
		\end{tabular}
	\vspace{-3mm}
\end{table}

\begin{table}[!t]\footnotesize  
	\centering
	\caption{{Comparison of different self-supervised strategies.}}\label{tabel:abla3}
	\setlength{\tabcolsep}{3.5pt}
	\begin{tabular}{l|c}
		\hline
		
		\hline
		Training Strategy &  PSNR$\uparrow$ / SSIM$\uparrow$ / LPIPS$\downarrow$  \\
		
		\hline
		SelfDRSC w/o self-distillation  &  28.411 / 0.8848 / 0.0574  \\
		SelfDRSC  &  28.704 / 0.8886 / 0.0546 \\
		SelfDRSC+  &  28.821 / 0.8902 / 0.0488 \\
		SelfDRSC++  &  {29.584} / \textbf{0.8989} / \textbf{0.0341}   \\
		SelfDRSC++ w/ self-distillation  &  \textbf{29.586} / \textbf{0.8989} / \textbf{0.0341}   \\
		\hline
		
		\hline	
	\end{tabular}
\vspace{-3mm}
\end{table}

For different VFI methods in $\mathcal{W}$, we use RIFE~\cite{huang2022rife}, AMT~\cite{li2023amt}, IFRNet~\cite{kong2022ifrnet} and EMA-VFI~\cite{zhang2023extracting} as $\mathcal{W}$, respectively.
Similarly, we use all these methods trained with the combined approach [D, R] as $\mathcal{W}$, which are also provided by~\cite{zhong2023clearer}. From Table~\ref{tabel:abla}, [D, R] RIFE achieves the best performance on self-supervised dual reversed rolling shutter correction task. Hence, we use [D, R] RIFE as our RS reconstruction module $\mathcal{W}$ in the following. 
{This indicates that $\mathcal W$ is not an arbitrary plug-in: its RS reconstruction fidelity directly affects the reliability of the self-supervised signal.}
Similarly, in Fig.~\ref{fig:VFI}, we conduct a visual comparison of RS reconstruction results obtained using different VFI methods. 
{This confirms that more faithful RS reconstruction by $\mathcal W$ provides stronger cycle-consistency supervision for training $\mathcal F$.}
{Although all compared VFI backbones use the same [D,R] setting, RIFE better matches the row-wise time-varying GS-to-RS synthesis in our framework, leading to higher-quality RS reconstruction.}
{We also finetuned RIFE on RS data to create a variant of RS reconstruction module $\mathcal{W}$ and the comparison of quantitative metrics (PSNR$\uparrow$ / SSIM$\uparrow$ / LPIPS$\downarrow$) is as follows: 29.863 (+0.279) / 0.9054 (+0.0065) / 0.0325 (-0.0016). Despite these gains, this approach violates the design principle of training without GS images. Moreover, we observe that the visual improvements on real-world data are not substantial.}
\begin{figure}[!t]\footnotesize
	\centering
	\includegraphics[width=0.5\linewidth]{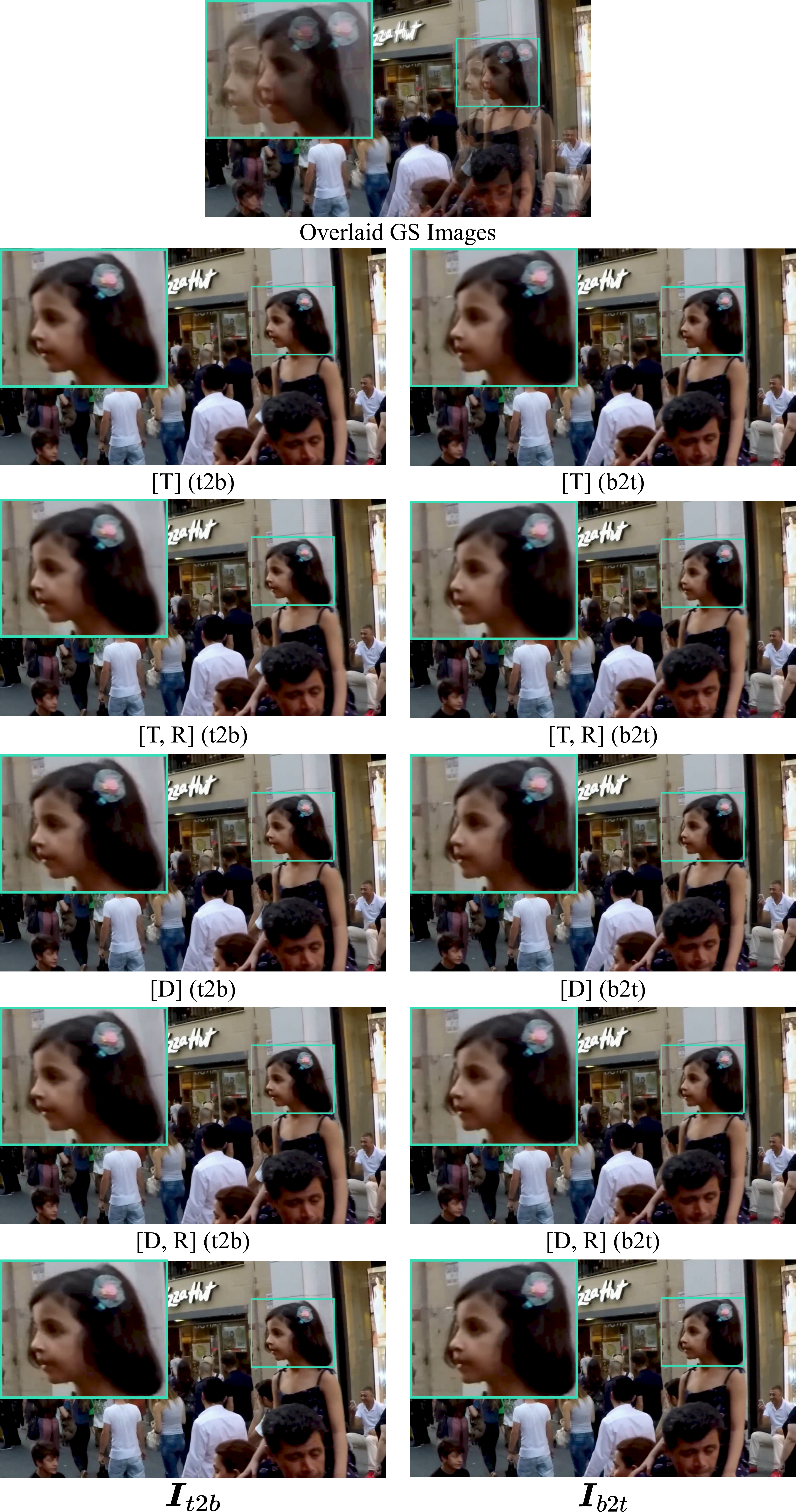}\\
	\vspace{-5mm}
	\caption{Visual comparison of RS reconstruction using different RIFE variants with GS frames at times $t_1$ and $t_H$. {[T], [D], and [R] denote the time index, distance index, and iterative reference-based estimation strategy in VFI, respectively.}
	}
\vspace{-3mm}
	\label{fig:varient}
\end{figure}
\begin{figure}[!t]\footnotesize
	\centering
	\includegraphics[width=0.5\linewidth]{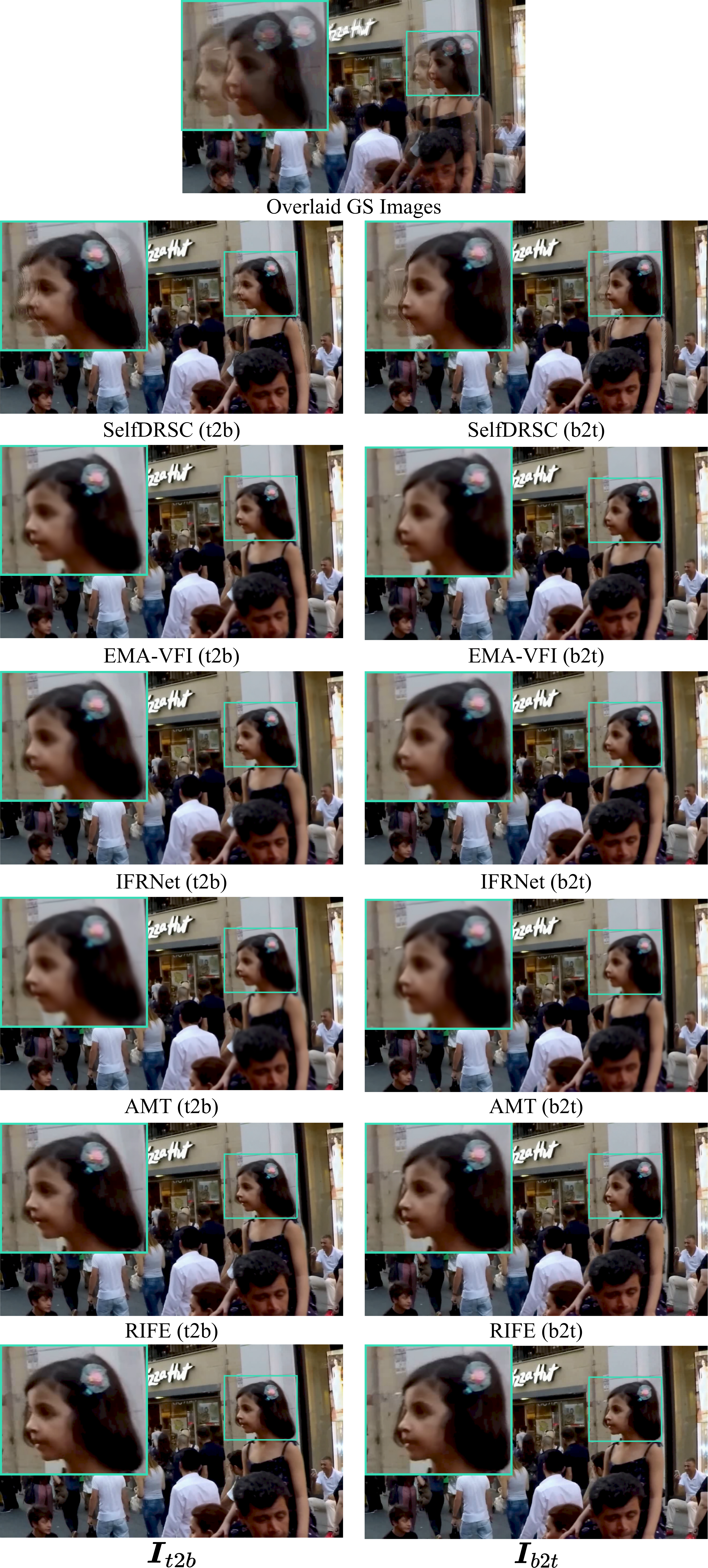}\\
	\vspace{-5mm}
	\caption{
		Visual comparison of RS reconstruction using different VFI methods with GS frames at times $t_1$ and $t_H$.
	}
\vspace{-3mm}
	\label{fig:VFI}
\end{figure}

To validate the role of the correlation matching module within $\mathcal{F}$ and to evaluate the impact of varying the number of fields {and the window size $w$ of the bidirectional correlation matching block}, we have created corresponding variants for these factors. These include a variant with the correlation matching module removed, along with the corresponding update block. Additionally, we have set {the number of fields to 1, 3, 5, 7 and the window size to 3, 5, 7 to investigate their impacts.}
From Table~\ref{tabel:abla2}, one can see that setting the number of fields as 3 gets the best performance and we use it as our default setting.
{For the window size of the bidirectional correlation matching block, the results indicate that while a larger window size generally leads to a higher performance, it also comes with a higher computational cost. The computational complexity for this part is $\mathcal{O}(w^2 \cdot H \cdot W)$. To balance computational load and performance, we ultimately chose a window size of 3.}

\begin{figure*}[!t]\footnotesize
	\centering
	\includegraphics[width=\linewidth]{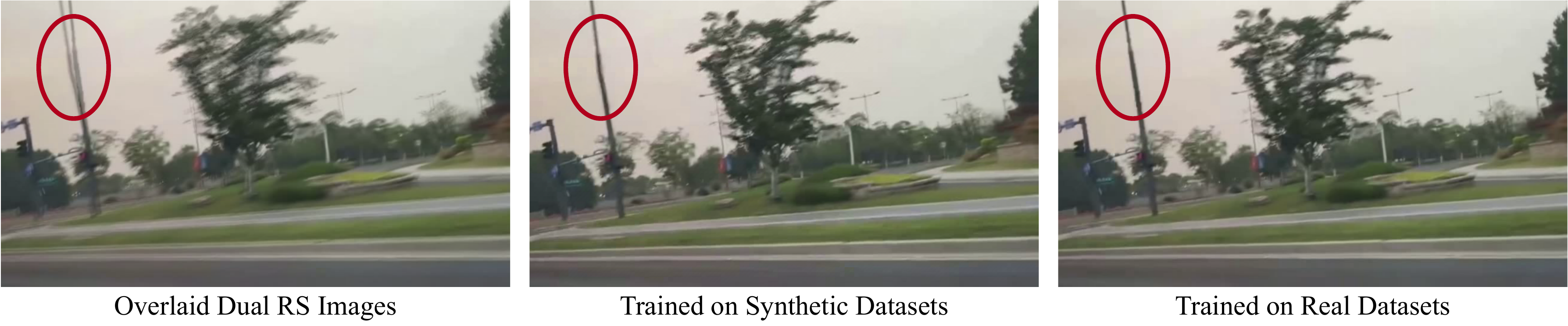}\\
	\vspace{-5mm}
	\caption{{Comparison of the real-data trained and synthetic-data trained models in correcting real RS distortion.}}
	\vspace{-3mm}
	\label{fig:real_vs_synthesis}
\end{figure*}
To further substantiate the effectiveness of our novel self-supervised strategy, we integrated the GS reconstruction module $\mathcal{W}$ (\ie, RIFE [D, R]) into SelfDRSC~\cite{shang2023self}, dubbing the enhanced model as SelfDRSC+. 
One can observe from the second row and the sixth row in Fig.~\ref{fig:VFI} that compared to SelfDRSC, our new self-supervised strategy generates more accurate RS reconstructions, which aids the correction module in conducting more effective self-supervised learning.
As demonstrated in Table~\ref{tabel:abla3}, SelfDRSC+ and SelfDRSC++ exhibit superior performance compared to the original SelfDRSC \cite{shang2023self}. More importantly, the proposed SelfDRSC++ simplifies the training process, transitioning from a two-stage approach to a single-stage one, eliminating the self-distillation phase while achieving improved performance.

{
	Additionally, we have also incorporated self-distillation (i.e., multi-stage training) into our new self-supervised strategy. However, we found that it hardly improves the performance of our method. This is because the self-distillation strategy in the previous version was mainly used to mitigate the boundary artifacts caused by warping-based RS reconstruction. Since our VFI-based RS reconstruction does not produce boundary artifacts, employing self-distillation does not further enhance the performance.
}

{To further demonstrate the effectiveness of our self-supervised strategy, we compare models trained with self-supervision on real data versus those trained on synthetic data. 
	The real dataset was split into 40 sequences for training and 10 sequences for testing. Both the test and training sets contain video segments from different readouts. The training settings remain consistent with those described in Section~\ref{sec: ID}.
	Visual comparisons are shown in Fig. \ref{fig:real_vs_synthesis}. In cases of severe pole distortion caused by high-speed camera motion, our model trained on real datasets still achieves good correction results. This further proves the effectiveness of our self-supervised strategy. For other types of RS cameras, our self-supervised strategy allows for direct retraining or finetuning of models without needing GT, thereby enhancing performance on corresponding dataset.} {We also replace the fixed-interval sampling of $t_m$ with uniform sampling over $[t_1,t_H]$ during training. The resulting model achieves $29.578/0.8987/0.0341$ in PSNR/SSIM/LPIPS on the standard 9-frame RS-GOPRO test protocol, nearly identical to the original $29.584/0.8989/0.0341$, indicating that SelfDRSC++ is insensitive to the sampling strategy of $t_m$.}

\begin{figure}[!t]\footnotesize
	\centering
	\includegraphics[width=\linewidth]{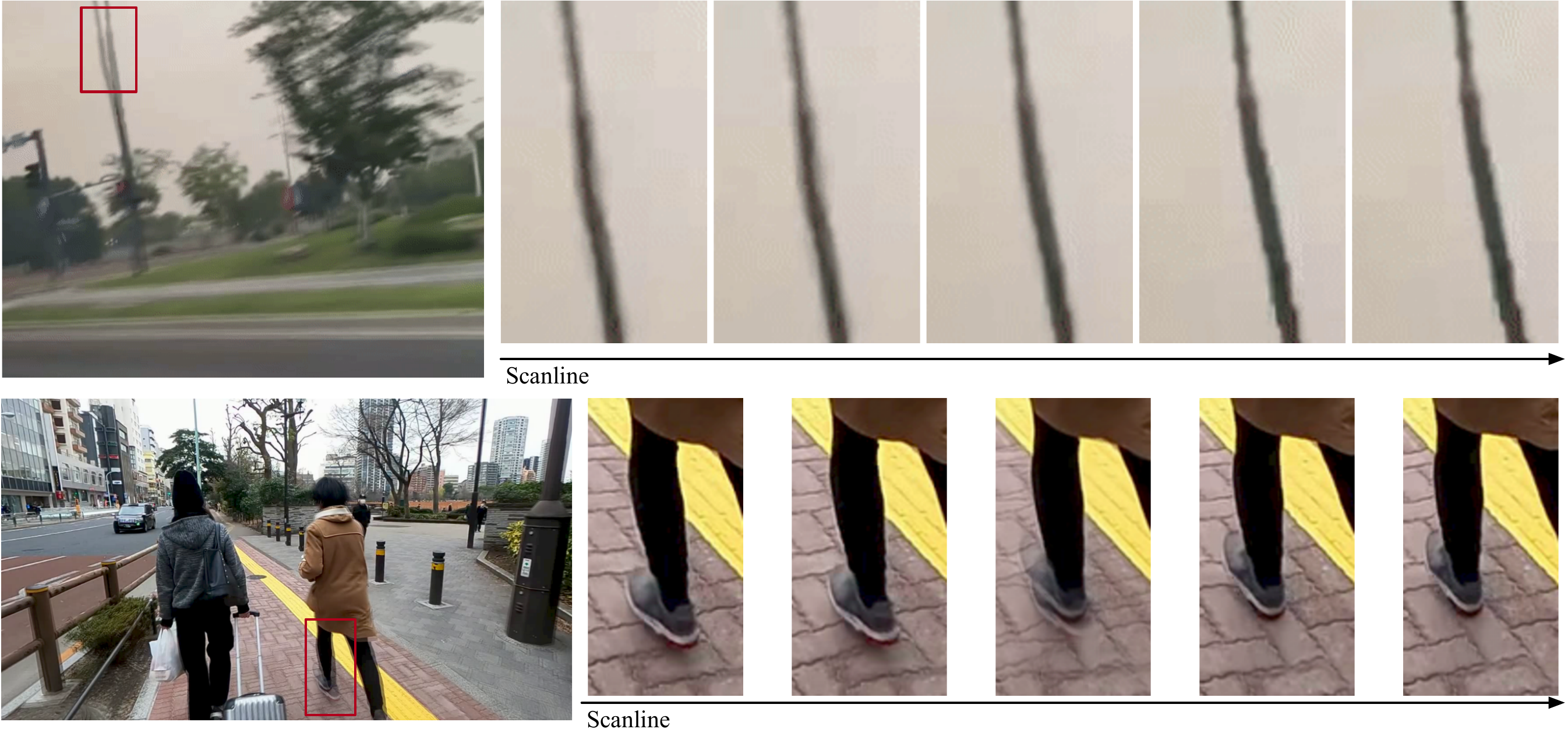}\\
	\vspace{-5mm}
	\caption{{Representative challenging cases under nonlinear motion.
	}
	}
\vspace{-3mm}
	\label{fig:failure}
\end{figure}
{
\section{Limitations and discussion}
Although SelfDRSC++ achieves strong performance on both synthetic and real-world data, several limitations remain. The proposed framework relies on row-wise temporal sampling and VFI-based reconstruction to model RS formation. When motion becomes highly nonlinear, such as extremely fast camera motion or non-rigid object motion, reconstruction errors may accumulate and reduce correction quality as shown in Fig.~\ref{fig:failure}.
In the first example, a streetlight pole captured from a fast-moving vehicle exhibits strong scanline-dependent deformation. In the second example, rapid leg motion of a running person introduces local temporal ambiguity. Although SelfDRSC++ can still alleviate most visible RS artifacts, residual deformation, blur, or local temporal discontinuities may remain in these extreme regions.

In addition, the current framework assumes synchronized and calibrated dual reversed RS inputs, and therefore cannot be directly extended to standard RS videos with large or irregular inter-frame intervals. Residual camera misalignment may further reduce reconstruction accuracy, since the network predicts latent GS frames conditioned on the observed input pair rather than explicitly correcting calibration errors. Although the synchronized beam-splitter setup largely suppresses inter-camera illumination inconsistency, extreme saturation or rapid illumination variation during the readout period may still weaken the photometric cycle-consistency signal. Furthermore, RS correction remains inherently ill-posed under heavy occlusion or highly irregular motion, where local ambiguities may still produce artifacts. Extending the framework to more general RS video settings may require longer-range temporal modeling, additional physical priors, or more flexible acquisition strategies.
}

\section{Conclusion} \label{sec:conclusion}
{In this paper, we propose SelfDRSC++, a self-supervised framework for dual reversed rolling shutter correction. {By formulating GS-to-RS reconstruction as a row-wise time-indexed VFI problem, the proposed method establishes a physically constrained RS--GS--RS cycle without requiring GS supervision.} We further introduce a lightweight DRSC network and a VFI-based RS reconstruction module guided by distorted time maps. Experimental results on synthetic and real-world datasets demonstrate that SelfDRSC++ achieves competitive or superior performance compared with existing supervised and self-supervised methods, while producing temporally coherent high-frame-rate GS videos.}

\section*{Funding}
This work was supported in part by the National Natural Science Foundation of China (62576241, 62172127, and U22B2035).

\section*{CRediT authorship contribution statement}
\textbf{Wei Shang:} Writing – original draft, Visualization, Validation, Methodology, Investigation, Software. \textbf{Dongwei Ren:} Writing – review \& editing, Resources, Supervision. \textbf{Wanying Zhang:} Visualization, Validation. \textbf{Qilong Wang:} Conceptualization. \textbf{Pengfei Zhu:} Conceptualization. \textbf{Wangmeng Zuo:} Resources, Supervision.

\section*{Declaration of competing interest}
The authors declare that they have no known competing financial interests or personal relationships that could have appeared to influence the work reported in this paper.

\section*{Data availability}

Data and software supporting this manuscript are available at \url{https://github.com/shangwei5/SelfDRSC_plusplus}.

\bibliographystyle{elsarticle-num} 
\bibliography{sn-bibliography}

\end{document}